\documentclass{article} 
\usepackage{iclr2027_conference,times}

\usepackage{amsmath,amsfonts,bm}

\def\eqref#1{equation~\ref{#1}}

\def\1{\bm{1}}

\DeclareMathAlphabet{\mathsfit}{\encodingdefault}{\sfdefault}{m}{sl}
\SetMathAlphabet{\mathsfit}{bold}{\encodingdefault}{\sfdefault}{bx}{n}

\usepackage{hyperref}
\usepackage{url}

\usepackage{graphicx}

\usepackage{booktabs} \usepackage{multirow} \usepackage{makecell}  
\usepackage{wrapfig}
\usepackage[table]{xcolor}
\usepackage{comment}

\usepackage[T1]{fontenc}
\usepackage[most]{tcolorbox}
\newtcblisting{promptbox}[1]{
  enhanced,
  breakable,
  listing only,
  title={#1},
  colback=black!2,
  colframe=black!45,
  colbacktitle=black!8,
  coltitle=black,
  fonttitle=\bfseries\small,
  boxrule=0.5pt,
  arc=1pt,
  left=5pt,
  right=5pt,
  top=5pt,
  bottom=5pt,
  listing options={
    basicstyle=\ttfamily\footnotesize,
    breaklines=true,
    columns=fullflexible,
    keepspaces=true,
    showstringspaces=false,
    upquote=true
  }
}

\title{From Prompting to Composing: A Spatial Canvas Interface for Poster Generation}

\author{\textbf{Yitong Wang}$^{1,2*}$\quad \textbf{Fangyun Wei}$^{2*}$\quad \textbf{Jinjing Zhao}$^{3,2}$\quad \textbf{Sirui Zhang}$^{4,2}$\\ \textbf{Hongyang Zhang}$^5$\quad \textbf{Dong Chen}$^2$\quad \textbf{Bo Dai}$^{6\dagger}$\quad \textbf{Yan Lu}$^2$ \\
$^1$Fudan University\quad $^2$Microsoft Research\quad $^3$The University of Sydney\\ $^4$USTC\quad $^5$University of Waterloo\quad $^6$The University of Hong Kong \\
$^*$Equal contribution.\quad $^\dagger$Corresponding author. \\
\small \mbox{\texttt{wangyitong23@m.fudan.edu.cn}}\quad \mbox{\texttt{\{fawe,doch,yanlu\}@microsoft.com}} \\ \small \mbox{\texttt{jzha0100@sydney.edu.au}}\quad \mbox{\texttt{zsr200901@mail.ustc.edu.cn}} \\ \small \mbox{\texttt{hongyang.zhang@uwaterloo.ca}}\quad \mbox{\texttt{bdai@hku.hk}} \\ \small Project page: \url{https://snowflakewang.github.io/Compo-Page/}
}

\iclrfinalcopy 
\begin{document}

\maketitle

\begin{figure}[h]
\begin{center}
     \includegraphics[width=\textwidth]{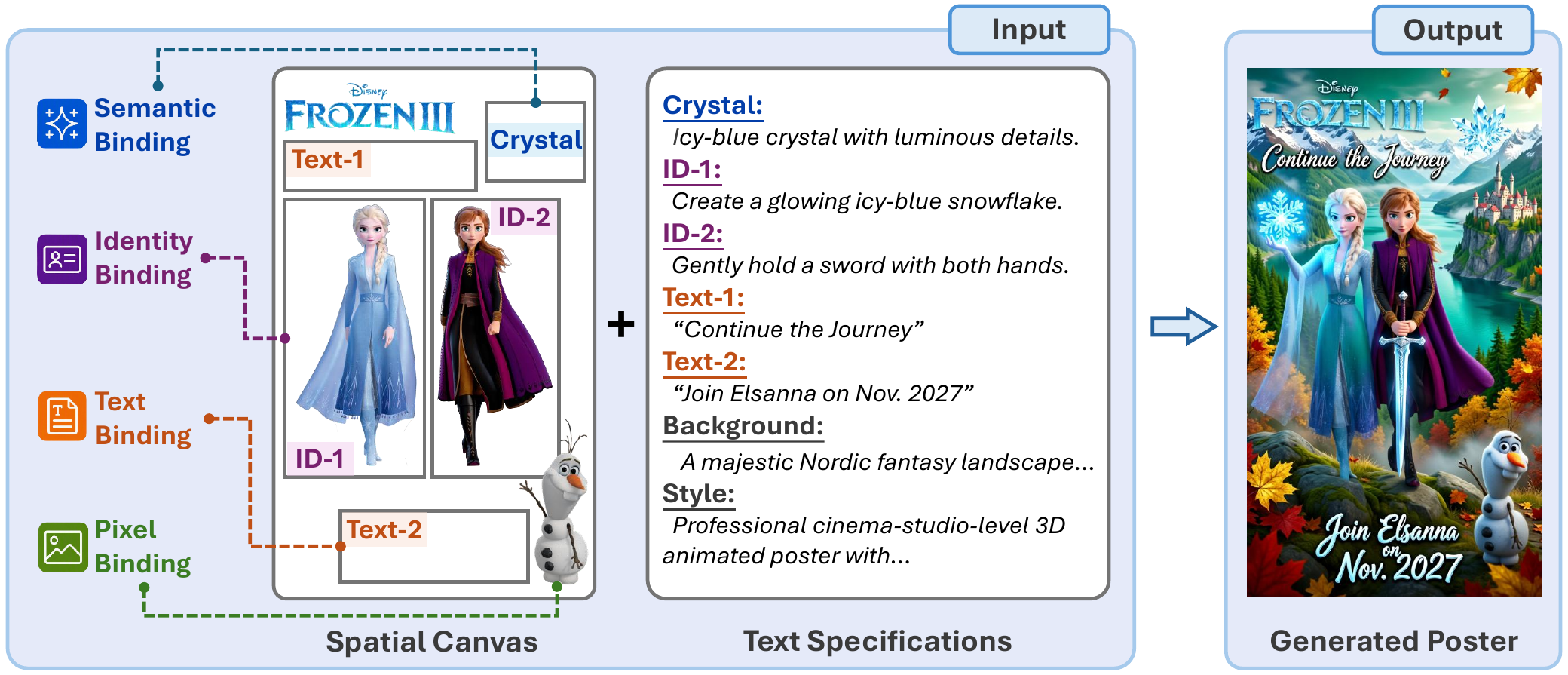}
\end{center}
\vspace{-4mm}
\caption{
Instead of describing an entire poster through a \textit{single text prompt}, our \textbf{Spatial Canvas Interface} allows users to directly compose generation intent in two-dimensional space.
The interface supports four complementary forms of binding:
\textcolor[RGB]{3,62,169}{\textbf{Semantic Binding}} specifies what visual content should be generated in a designated region;
\textcolor[RGB]{120,32,110}{\textbf{Identity Binding}} associates a reference image with a target region to preserve its visual identity;
\textcolor[RGB]{192,79,21}{\textbf{Text Binding}} places specified text at a desired location; and
\textcolor[RGB]{59,125,35}{\textbf{Pixel Binding}} directly anchors visual content that should be preserved.
The accompanying \textit{Text Specifications} further provide \textit{detailed instructions} for individual bindings, such as the desired appearance, action, or content of each bound element, together with global specifications for the background and style.
By jointly expressing spatial structure on the canvas and detailed generation requirements in text, our interface provides an explicit and structured specification of the target poster, enabling generation through composition rather than text description alone.
}
\label{fig:teaser}
\end{figure}

\begin{abstract}
Text prompting is an indirect interface for poster generation, requiring users to encode inherently two-dimensional composition intent into a one-dimensional sequence of words.
We introduce a \textbf{Spatial Canvas Interface} that enables users to directly compose generation intent in space through four complementary binding types: semantic, identity, text, and pixel, together with Text Specifications for individual elements and global appearance.
Based on this interface, we develop \textbf{Compo}, a poster generation model adapted from a pretrained image editing model to understand Spatial Canvas inputs and Text Specifications.
Compo supports both direct inference, where users explicitly construct the canvas, and agentic mode, where a high-level request is automatically translated into a planned Spatial Canvas.
To train Compo, we develop a scalable pipeline that automatically constructs supervision data for different binding types and their combinations, enabling efficient adaptation without training a specialized poster generator from scratch.
We further introduce a benchmark that evaluates adherence to individual binding types and their joint composition.
Experiments show that Compo achieves stronger compositional controllability than both general-purpose image generation models and dedicated poster generation systems while maintaining high visual quality.
By decoupling intent specification from visual generation, our work shifts poster generation from \emph{prompting} toward \emph{composing}.
\end{abstract}

\section{Introduction}
\vspace{-2mm}

Text has become the direct interface for modern generative models, including recent systems for poster generation~\citep{chen2025posta, chen2025postercraft}.
Given a prompt, a model is expected to translate a one-dimensional sequence of words into a complete two-dimensional visual composition.
While recent image generation models~\citep{wu2025qwenimage, cai2025z, liu2026ernie, team2025longcatimage} can produce visually compelling posters, text alone remains a limited interface for specifying \emph{how a poster should be composed}.
Posters are inherently structured, with subjects, text, logos, and graphical elements occupying intentional locations and often requiring specific identities or visual content to be preserved.
Encoding such spatial and visual constraints entirely through language is indirect and ambiguous, especially as compositions become more complex.

We therefore explore a different paradigm: \textit{instead of describing the entire poster with text, users compose generation intent directly in space}.
As illustrated in Figure~\ref{fig:teaser}, we introduce a \textbf{Spatial Canvas Interface} for \textit{poster generation} that decouples intent specification from visual generation.
The Spatial Canvas represents where different elements should appear and how they should be bound, while accompanying \textbf{Text Specifications} provide detailed instructions for individual elements as well as global descriptions such as background and style.
Together, they form a structured representation of the target poster.

Our interface supports four complementary forms of binding.
\textit{Semantic Binding} associates a textual concept with a target region;
\textit{Identity Binding} associates a reference image with a region to preserve its visual identity;
\textit{Text Binding} specifies both an exact text string and its desired location; and
\textit{Pixel Binding} directly places visual content that should be preserved.
These bindings can be freely combined within a single canvas, providing unified control over semantics, identity, text, pixels, and layout.
From this perspective, layout control~\citep{zheng2023layoutdiffusion, zhang2025creatilayout}, reference-driven generation~\citep{xia2025dreamomni2, wu2026omnigen2}, text rendering~\citep{chen2024textdiffuser2, peng2025bizgen}, and image editing~\citep{wu2025qwenimage, team2025longcatimage} can all be viewed as instances of a common principle: \textit{binding generation constraints to spatial locations on a shared canvas}.

To realize this interface, we develop \textbf{Compo}, a poster generation model adapted from a pretrained image editing model to understand Spatial Canvas inputs and their associated Text Specifications.
Compo supports two complementary inference modes.
In \textit{direct inference}, users explicitly construct the Spatial Canvas and Text Specifications for fine-grained control.
In \textit{agentic mode}, users provide only a high-level natural-language request, and an agent automatically converts it into the same structured representation by extracting visual elements and text, planning their spatial layout, selecting appropriate binding types, and retrieving reference images when needed.
The resulting Spatial Canvas and Text Specifications are then consumed by Compo to generate the final poster.
In this way, the Spatial Canvas serves both as a human-facing composition interface and as a structured intermediate representation for agentic generation.

Training Compo presents a practical challenge, as existing image generation and editing models are not designed to interpret such heterogeneous spatial bindings, while collecting large-scale, human-designed canvas--poster pairs would be prohibitively expensive. To address this challenge, we develop a \textit{scalable training data generation pipeline} that automatically constructs supervision for individual binding types and their combinations, resulting in \textbf{Compo-200K}. Using this automatically generated dataset, we efficiently adapt a pretrained image editing model to the Spatial Canvas Interface, without training a specialized poster generation model from scratch.

Finally, we introduce a \textbf{benchmark and evaluation protocol} for spatially compositional poster generation.
Rather than evaluating only overall image quality, our benchmark measures adherence to each binding type individually, as well as their joint composition.
Experiments show that Compo achieves stronger controllability than both general text-and-image-to-image generation models, such as Qwen-Image-Edit-2511~\citep{wu2025qwenimage}, and dedicated poster generation systems, such as CreatiDesign~\citep{zhang2026creatidesign}, while maintaining high visual quality.
\vspace{-2mm}
\section{Related Works}
\vspace{-3mm}

\textbf{Image Generation and Editing Foundation Models.}
Recent commercial foundation models~\citep{maiimage, grokimagine, seedream, gptimage2, nanobanana2, qwenimage3} support both text-to-image generation and image editing, with strong visual quality, instruction following, and text rendering. Design-oriented models~\citep{ideogram-4-2026, reve} further introduce structured text or image inputs for more explicit control over layout and composition. Open-source models~\citep{wu2025qwenimage, team2025longcatimage, team2026firered, song2026joyai, flux-2-2025} commonly extend text-to-image generation to editing by combining multimodal language models~\citep{bai2025qwen25vltechnicalreport, bai2025qwen3vl} with diffusion transformers~\citep{peebles2023scalable, esser2024scaling}. Other efforts improve training efficiency through compact designs~\citep{chen2026lens, hu2026swift} or develop unified architectures~\citep{cai2026hidream, diao2026sensenova} that support generation, editing, and, in some cases, image understanding. Despite these advances, jointly expressing multiple multimodal references and precise layout constraints remains underexplored for complex graphic design.

\textbf{Reference- and Layout-Driven Image Generation.}
Controllable image generation methods can be broadly grouped by their use of reference and layout conditions. Reference-driven methods~\citep{ruiz2023dreambooth, wang2025ms, wu2025uno, tan2025ominicontrol, zhu2025multibooth, mou2025dreamo, cheng2025umo, she2026mosaic, garibi2025tokenverse, chen2026xverse, xia2025dreamomni2, wu2026omnigen2, xu2026unicustom, chen2026macro} preserve visual attributes such as subject identity, appearance, or style from user-provided images, while layout-driven methods~\citep{li2023gligen, wang2024instancediffusion, zhou2024migc, cheng2024hico, zhou2025dreamrenderer, zheng2023layoutdiffusion} control spatial composition through explicit structural constraints. Joint-control methods~\citep{qin2023unicontrol, wang2025unicombine, xu2026contextgen} combine multiple references and layouts, but often rely on separate or task-specific modules, limiting extensibility. 
Unified-input methods~\citep{dalva2026canvas, yi2026flowinone} instead encode heterogeneous control signals within a single input image. However, existing works mainly focus on human-centric portrait or general image synthesis, leaving structured graphic design underexplored. Furthermore, these approaches lack explicit and fine-grained bindings between spatial regions and their corresponding text descriptions. These limitations motivate a more general and structured interface for multimodal binding and layout control.

\textbf{Poster Generation.}
Poster generation requires joint modeling of textual content, graphic layout, visual assets, and design aesthetics. Early methods focus on individual aspects, including layout generation~\citep{hsu2023posterlayout, lin2023autoposter, seol2024posterllama, yang2024posterllava, chen2025posta} and text rendering~\citep{tuo2024anytext2, chen2024textdiffuser2, liu2024glyph, wang2025designdiffusion}. Later approaches~\citep{pu2025art, peng2025bizgen, gao2025postermaker} incorporate visual references and layout conditions for improved controllability, while others~\citep{hu2025dreamposter, zhang2025creatiposter} condition on background images or separately model foreground and background elements. Recent methods further improve aesthetic quality through reinforcement learning~\citep{chen2025postercraft, chen2026posteromni} or support multiple conditions~\citep{zhang2026creatidesign}. However, existing approaches still rely on restricted reference types or task-specific control modules, limiting extensibility and requiring complex user inputs.
\vspace{-3mm}
\section{Method}
\vspace{-2mm}

\textbf{Problem Formulation.}
Given a \textit{Spatial Canvas} and its accompanying \textit{Text Specifications}, our goal is to generate a poster that faithfully satisfies the specified spatial layout, semantic content, visual identities, textual elements, and preserved pixels, while maintaining high visual quality.
The \textit{Spatial Canvas} serves as the \textit{visual modality}, encoding element locations, reference assets, preserved visual content, and binding types in two-dimensional space.
The \textit{Text Specifications} serve as the \textit{language modality}, providing detailed element-level instructions together with global descriptions such as background and style.
Together, these two modalities form the input to our model, \textbf{Compo}.

We describe our scalable training data generation pipeline in Section~\ref{sec:data}.
Section~\ref{sec:architecture} presents the architecture and the two-stage training procedure, including supervised training and reinforcement learning.
Finally, Section~\ref{sec:inference} introduces the inference pipeline, covering both direct inference and agentic generation.

\begin{figure}[t]
\begin{center}
\includegraphics[width=\textwidth]{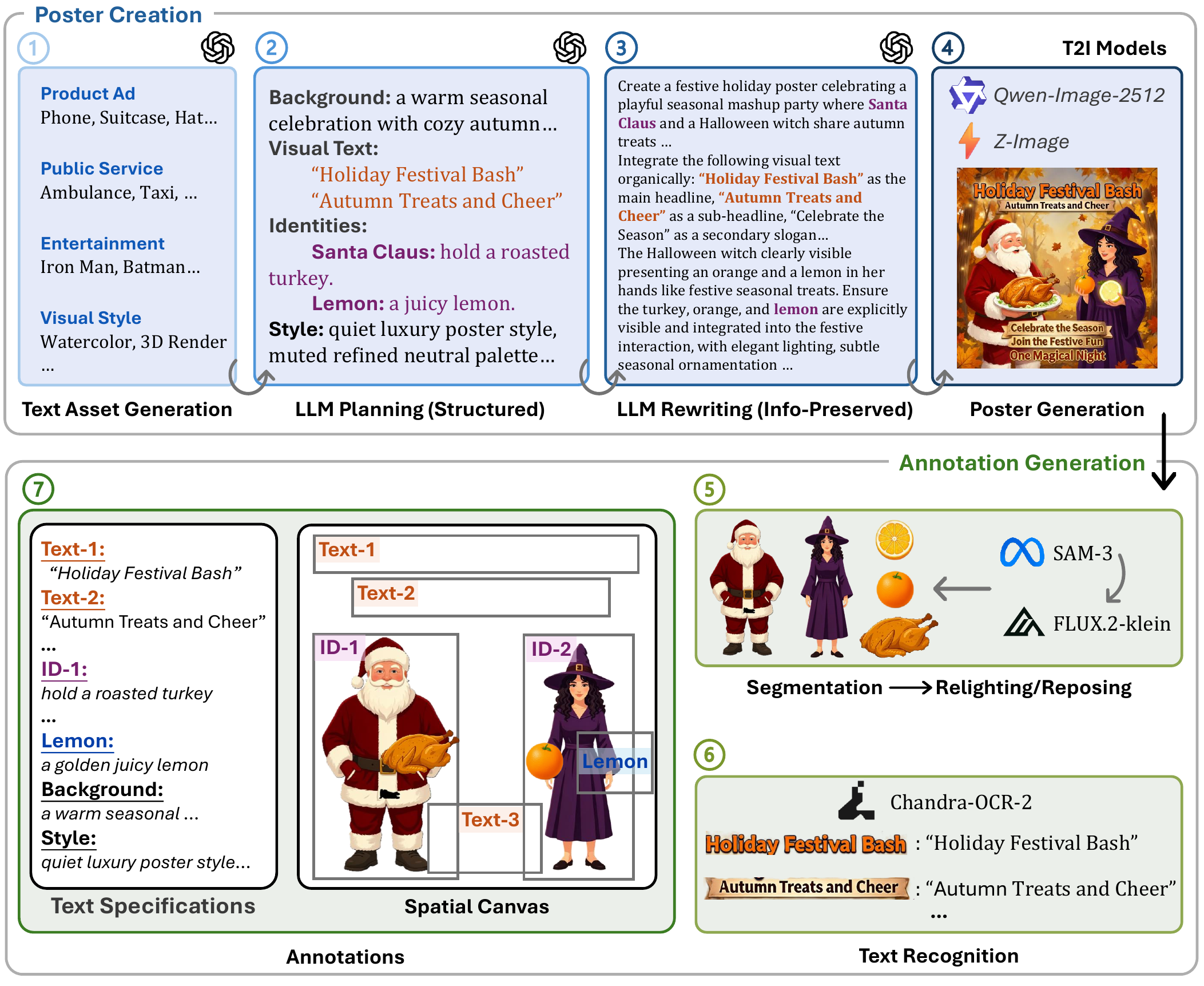}
\end{center}
\vspace{-6mm}
\caption{
\textbf{Training data generation pipeline.}
We first create diverse posters through structured planning and information-preserving rewriting with \textit{GPT-5.5}~\citep{gpt55}, followed by rendering with \textit{Qwen-Image-2512}~\citep{wu2025qwenimage}, \textit{Z-Image}~\citep{cai2025z}, and \textit{ERNIE-Image}~\citep{liu2026ernie}.
We then automatically construct paired Spatial Canvases and Text Specifications through instance segmentation, visual asset processing, and text recognition, providing scalable supervision for training Compo.
}
\vspace{-3mm}
\label{fig:data_pipeline}
\end{figure}

\vspace{-3mm}
\subsection{Training Data Generation}
\label{sec:data}
\vspace{-2mm}

Training Compo requires paired supervision comprising a target poster, its Spatial Canvas, and corresponding Text Specifications. However, existing poster datasets~\citep{chen2025postercraft, mao2026textground4m} lack the diversity and fine-grained annotations required by our interface, while manual construction at a large scale is prohibitively expensive. Inspired by recent works~\citep{zhang2026creatidesign, chen2026posteromni}, we develop a scalable data generation pipeline that automatically generates posters with fine-grained multimodal bindings and layout annotations. As illustrated in Figure~\ref{fig:data_pipeline}, our pipeline consists of two stages: \textbf{poster creation} and \textbf{annotation generation}.

\textbf{Poster Creation.} We use \textit{GPT-5.5}~\citep{gpt55}, \textit{Claude Opus 4.6}~\citep{claudeopus46}, and \textit{Gemini 3.5 Flash}~\citep{gemini35flash} to construct two text asset repositories: a content repository containing diverse poster themes and their associated items, spanning categories such as product advertising, public service, and entertainment, and a style repository covering a wide range of visual styles. To construct diverse poster concepts, we sample a theme with its associated items and a visual style from the respective repositories. GPT-5.5 then organizes these sampled assets into a structured specification and enriches it by planning the visual text and poster background, as well as detailed descriptions and interactions for the identities or objects instantiated from the sampled items. The resulting specification is rewritten into a natural poster generation prompt while preserving all sampled and planned information. Finally, we render the prompts with multiple text-to-image models, including \textit{Qwen-Image-2512}~\citep{wu2025qwenimage}, \textit{Z-Image}~\citep{cai2025z}, and \textit{ERNIE-Image}~\citep{liu2026ernie}, selected for their advanced text rendering capabilities, to enhance the diversity of visual appearance and composition in the training data.

\textbf{Annotation Generation.}
Given a generated poster, we automatically recover the structured supervision required by the Spatial Canvas Interface.
For visual elements, we use \textit{SAM-3}~\citep{carion2026sam3} to segment relevant instances from the generated poster.
The segmented assets are further processed with \textit{FLUX.2-klein-9B}~\citep{flux-2-2025} for relighting or reposing when needed, producing suitable visual references for constructing different binding types.
For textual elements, we use \textit{Chandra-OCR-2}~\citep{chandra2} to recognize the rendered text and recover its spatial location.
As shown in Figure~\ref{fig:data_pipeline}, all elements, including bounding boxes, reference images, and textual identifiers, are directly rendered onto the Spatial Canvas as visual conditions. In parallel, we construct the Text Specifications from the original structured planning information to record element-level descriptions and global attributes such as background and style. Consequently, each training sample consists of a generated poster paired with its Spatial Canvas and Text Specifications, providing scalable supervision for training Compo.
As a result, we construct a new dataset of roughly 200K samples with fine-grained structured annotations, denoted as \textbf{Compo-200K}. More details about data creation and annotation are included in Appendix~\ref{sec:dataset-details}.

\begin{figure}[!t]
\begin{center}
    \includegraphics[width=\textwidth]{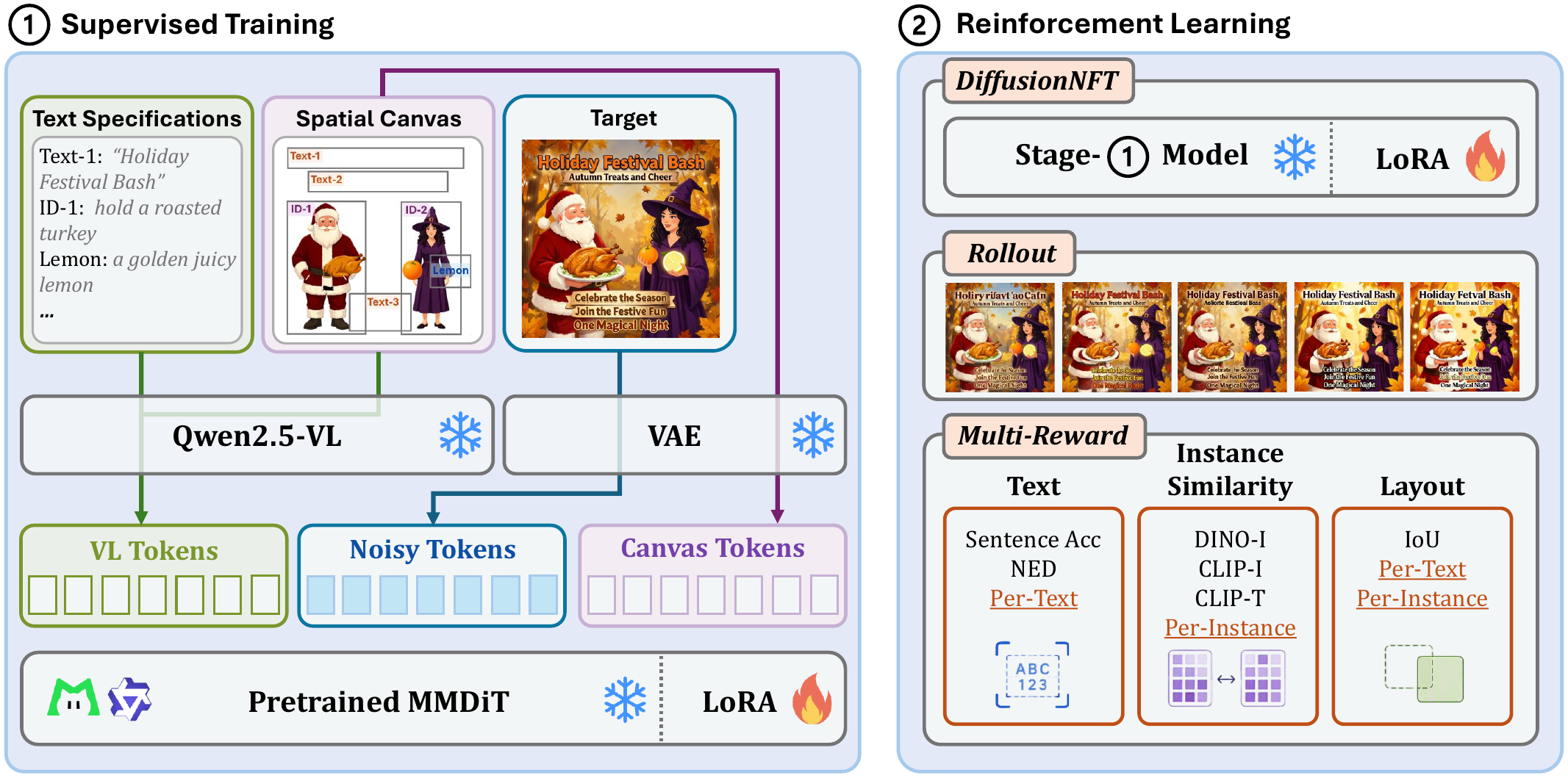}
\end{center}
\vspace{-6mm}
\caption{
\textbf{Architecture and training pipeline of Compo.}
Compo is trained in two stages.
During supervised training, the Spatial Canvas and Text Specifications are jointly encoded with the target poster to adapt a pretrained MMDiT~\citep{peebles2023scalable, esser2024scaling} through LoRA~\citep{hu2021lora}.
We then apply DiffusionNFT~\citep{zheng2026diffusionnft} for reinforcement learning with complementary rewards for text accuracy, instance similarity, and layout adherence.
}
\vspace{-4mm}
\label{fig:model}
\end{figure}

\vspace{-3mm}
\subsection{Architecture}
\label{sec:architecture}
\vspace{-2mm}

As illustrated in Figure~\ref{fig:model}, Compo is built upon a pretrained multimodal diffusion transformer (MMDiT)~\citep{peebles2023scalable, esser2024scaling} and trained in two stages: \textbf{Supervised Training} and \textbf{Reinforcement Learning}.
Rather than modifying the backbone architecture, we adapt the pretrained model using LoRA~\citep{hu2021lora}, allowing Compo to efficiently learn the proposed Spatial Canvas Interface while retaining the generative capabilities of the pretrained model.

\textbf{Stage-1: Supervised Training.}
We first perform supervised training using the data generated by the pipeline described in Section~\ref{sec:data}.
Each training sample consists of a target poster together with its corresponding \textit{Spatial Canvas} and \textit{Text Specifications}.
We experiment with both \textit{LongCat-Image-Edit}~\citep{team2025longcatimage} and \textit{Qwen-Image-Edit-2511}~\citep{wu2025qwenimage} as pretrained image editing backbones, and follow the same training objective and loss formulation as their respective original models.
Both models use a frozen Qwen2.5-VL~\citep{bai2025qwen25vltechnicalreport} to encode the multimodal specifications into visual-language tokens (VL tokens).
The Spatial Canvas and target poster are encoded into the latent space using a frozen VAE, producing canvas tokens and target latent tokens, respectively.
During diffusion training, noise is added to the target latents, and the pretrained MMDiT jointly processes the noisy target tokens, canvas tokens, and VL tokens. Specifically, the text branch of MMDiT processes the VL tokens, while the image branch processes the noisy target tokens together with the canvas tokens.
The pretrained components are kept frozen, while LoRA parameters attached to the MMDiT are optimized.
This stage enables Compo to interpret heterogeneous bindings and generate posters that follow the structured Spatial Canvas and associated Text Specifications. More details are described in Appendix~\ref{sec:supervised-training-details}.

\begin{figure}[!t]
\begin{center}
    \includegraphics[width=\textwidth]{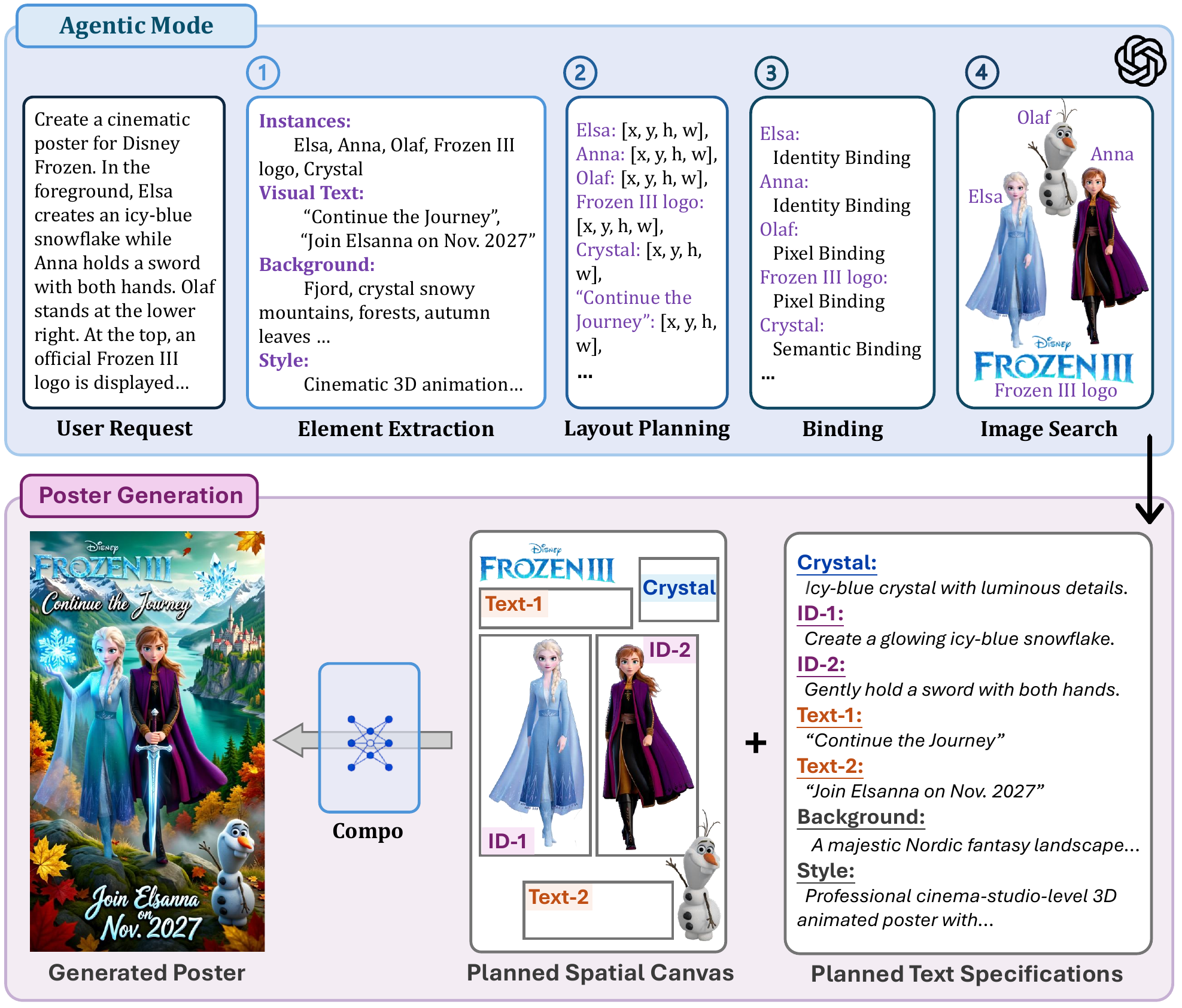}
\end{center}
\vspace{-6mm}
\caption{
\textbf{Poster generation using Compo with Agentic Mode.}
Given a high-level user request, a GPT-5.6-based agent extracts visual elements and text, plans their spatial layout, assigns appropriate binding types, and retrieves reference images when needed.
It then constructs a planned Spatial Canvas and Text Specifications, which are passed to Compo to generate the final poster.
}
\vspace{-5mm}
\label{fig:inference}
\end{figure}

\textbf{Stage-2: Reinforcement Learning (RL).}
While supervised training teaches Compo to follow the Spatial Canvas Interface, it does not explicitly optimize the fidelity of each individual binding.
We therefore further refine the supervised model using \textit{DiffusionNFT}~\citep{zheng2026diffusionnft}.
Starting from the Stage-1 model, we perform multiple rollouts for each input and optimize a trainable LoRA with a multi-reward objective tailored to the binding requirements of our interface.
We follow the optimization strategy and training hyperparameters of DiffusionNFT, while applying our binding-aware multi-reward objective for RL post-training.

Our reward comprises three complementary components. The \textbf{text reward} evaluates the rendering quality of each Text Binding using sentence accuracy and normalized edit distance (NED; i.e., 1 minus the edit distance)~\citep{gao2025postermaker}, capturing exact correctness and character-level fidelity, respectively. The \textbf{instance similarity reward} evaluates the consistency of each generated instance with its corresponding visual or semantic condition by combining DINO-I~\citep{oquab2023dinov2}, CLIP-I~\citep{radford2021clip}, and CLIP-T~\citep{radford2021clip} similarities, thereby encouraging both visual fidelity and semantic alignment. The \textbf{layout reward} measures the IoU between the intended and generated regions of each text or visual element, encouraging accurate spatial placement. For each rollout candidate, these rewards are first computed at the individual text or instance level and then aggregated into image-level scores. Sentence accuracy is aggregated using a logical AND operation, such that the image-level score is one only if all bound text elements are rendered correctly and zero otherwise. All other reward metrics are aggregated using arithmetic averaging. The final reward combines the resulting image-level signals to jointly improve text rendering, instance fidelity, and layout adherence. Further implementation details are provided in Appendix~\ref{sec:RL-details}.

\vspace{-2mm}
\subsection{Inference}
\label{sec:inference}
\vspace{-2mm}
Compo supports two complementary inference modes: \textit{Direct Inference} and \textit{Agentic Mode}.

\textbf{Direct Inference.}
In direct inference, users manually specify the Spatial Canvas and Text Specifications, providing explicit control over element locations, binding types, visual references, and generation requirements.

\textbf{Agentic Mode.}
As illustrated in Figure~\ref{fig:inference}, Agentic Mode allows users to generate a poster directly from a high-level natural-language request.
The agent first performs \textit{element extraction} to identify visual instances, visual text, background, and style.
It then performs \textit{layout planning} to determine the spatial location and size of each element, followed by \textit{binding selection} to assign an appropriate binding type.
For elements requiring visual references, the agent further performs \textit{image search} to retrieve suitable assets.
Based on these intermediate results, the agent constructs a planned Spatial Canvas and detailed Text Specifications, which are passed to Compo for final poster generation.
The detailed implementation of Agentic Mode is described in Appendix~\ref{sec:agentic-mode}.
\vspace{-2mm}
\section{Experiments}
\label{sec:experiments}
\vspace{-1mm}

\definecolor{bestcolor}{HTML}{FCE4D6}
\definecolor{secondcolor}{HTML}{D9EAF7}
\newcommand{\orangebox}[1]{%
  \begingroup
  \setlength{\fboxsep}{1pt}%
  \colorbox{bestcolor}{#1}%
  \endgroup
}

\newcommand{\bluebox}[1]{%
  \begingroup
  \setlength{\fboxsep}{1pt}%
  \colorbox{secondcolor}{#1}%
  \endgroup
}
\begin{table}[t]
\caption{We compare general text-and-image-to-image models, dedicated poster generation models, and our Compo variants across pixel, identity, semantic, and text binding metrics. Higher values indicate better performance. The \orangebox{best} and \bluebox{second-best} results are highlighted in orange and blue, respectively. ``NED'': normalized edit distance. ``Sen. Acc.'': sentence accuracy.}
\label{tab:benchmark}
\centering
\begingroup
\setlength{\tabcolsep}{2.3pt}
\renewcommand{\arraystretch}{1.12}
\resizebox{\linewidth}{!}{%
\begin{tabular}{@{}l*{17}{c}@{}}
\toprule
\multirow{2}{*}{\bfseries \shortstack[l]{Models}} & \multicolumn{5}{c}{\bfseries Pixel Binding}
& \multicolumn{5}{c}{\bfseries Identity Binding}
& \multicolumn{2}{c}{\bfseries Semantic Binding}
& \multicolumn{3}{c}{\bfseries Text Binding}
& \multicolumn{2}{c}{\bfseries Overall} \\
\cmidrule(lr){2-6}\cmidrule(lr){7-11}\cmidrule(lr){12-13}\cmidrule(lr){14-16}\cmidrule(lr){17-18}
& DINO-I & CLIP-I & M-DINO & SSIM & IoU
& DINO-I & CLIP-I & M-DINO & CLIP-T & IoU
& CLIP-T & IoU
& NED & \shortstack{Sen. Acc.} & IoU
& IoU & \shortstack{Avg. Score} \\
\midrule
\multicolumn{18}{l}{\textbf{\textit{General Text-and-Image-to-Image Models}}} \\
\addlinespace[1pt]
Step1X-Edit~\citep{liu2025step1x}
& 0.426 & 0.481 & 0.327 & 0.493 & 0.473
& 0.600 & 0.645 & 0.488 & 0.167 & 0.563
& 0.012 & 0.056
& 0.060 & 0.020 & 0.270
& 0.317 & 0.339 \\
\addlinespace[1pt]
GLM-Image~\citep{glmimage}
& 0.535 & 0.580 & 0.441 & 0.591 & 0.571
& 0.663 & 0.699 & 0.554 & 0.185 & 0.606
& 0.019 & 0.110
& 0.030 & 0.012 & 0.230
& 0.328 & 0.389 \\
\addlinespace[1pt]
BAGEL~\citep{deng2025bagel}
& 0.274 & 0.311 & 0.207 & 0.321 & 0.321
& 0.534 & 0.580 & 0.428 & 0.156 & 0.511
& 0.006 & 0.045
& 0.007 & 0.000 & 0.184
& 0.228 & 0.259 \\
\addlinespace[1pt]
ChronoEdit-14B~\citep{wu2025chronoedit}
& 0.575 & 0.629 & 0.479 & 0.647 & 0.638
& 0.697 & 0.733 & 0.596 & 0.192 & 0.631
& 0.011 & 0.055
& 0.004 & 0.000 & 0.260
& 0.350 & 0.410 \\
\addlinespace[1pt]
\shortstack[l]{FLUX.2-klein-9B}~\citep{flux-2-2025}
& 0.341 & 0.378 & 0.254 & 0.392 & 0.390
& 0.543 & 0.586 & 0.430 & 0.168 & 0.511
& 0.058 & 0.225
& 0.187 & 0.088 & 0.431
& 0.388 & 0.332 \\
\addlinespace[1pt]
\shortstack[l]{JoyAI-Image-Edit}~\citep{song2026joyai}
& 0.440 & 0.477 & 0.328 & 0.480 & 0.467
& 0.626 & 0.658 & 0.495 & 0.183 & 0.585
& 0.071 & 0.244
& 0.217 & 0.141 & 0.372
& 0.394 & 0.386 \\
\addlinespace[1pt]
\shortstack[l]{FireRed-Image-Edit-1.1}~\citep{team2026firered}
& 0.387 & 0.419 & 0.286 & 0.429 & 0.436
& 0.605 & 0.642 & 0.472 & 0.182 & 0.558
& 0.061 & 0.228
& 0.125 & 0.056 & 0.361
& 0.376 & 0.350 \\
\addlinespace[1pt]
\shortstack[l]{Boogu-Image-0.1-Edit}~\citep{chen2026boogu}
& 0.473 & 0.506 & 0.376 & 0.504 & 0.494
& 0.668 & 0.701 & 0.539 & 0.193 & 0.607
& 0.057 & 0.202
& 0.230 & 0.119 & 0.347
& 0.385 & 0.401 \\
\addlinespace[1pt]
HiDream-O1-Image~\citep{cai2026hidream}
& 0.624 & 0.669 & 0.514 & 0.685 & 0.663
& 0.747 & 0.776 & 0.642 & 0.218 & 0.672
& 0.138 & 0.415
& 0.183 & 0.102 & 0.512
& 0.547 & 0.504 \\
\addlinespace[1pt]
\shortstack[l]{SenseNova-U1}~\citep{diao2026sensenova}
& 0.099 & 0.120 & 0.076 & 0.123 & 0.170
& 0.249 & 0.306 & 0.179 & 0.096 & 0.304
& 0.038 & 0.215
& 0.078 & 0.055 & 0.232
& 0.218 & 0.156 \\
\addlinespace[1pt]
DreamOmni2~\citep{xia2025dreamomni2}
& 0.405 & 0.441 & 0.322 & 0.451 & 0.455
& 0.576 & 0.605 & 0.471 & 0.160 & 0.547
& 0.017 & 0.085
& 0.053 & 0.020 & 0.274
& 0.310 & 0.326 \\
\addlinespace[1pt]
OmniGen2~\citep{wu2026omnigen2}
& 0.345 & 0.380 & 0.255 & 0.384 & 0.376
& 0.517 & 0.552 & 0.401 & 0.149 & 0.506
& 0.018 & 0.100
& 0.015 & 0.001 & 0.227
& 0.267 & 0.282 \\
\addlinespace[1pt]
LongCat-Image-Edit~\citep{team2025longcatimage}
& 0.515 & 0.562 & 0.406 & 0.581 & 0.574
& 0.723 & 0.775 & 0.620 & 0.221 & 0.664
& 0.072 & 0.232
& 0.150 & 0.055 & 0.409
& 0.445 & 0.437 \\
\addlinespace[1pt]
\shortstack[l]{Qwen-Image-Edit-2511}~\citep{wu2025qwenimage}
& 0.567 & 0.608 & 0.440 & 0.624 & 0.610
& 0.720 & 0.752 & 0.606 & 0.210 & 0.653
& 0.049 & 0.190
& 0.133 & 0.072 & 0.399
& 0.434 & 0.442 \\
\midrule
\multicolumn{18}{l}{\textbf{\textit{Dedicated Poster Generation Models}}} \\
\addlinespace[1pt]
PosterMaker~\citep{gao2025postermaker}
& 0.708 & 0.748 & 0.611 & 0.766 & 0.760
& 0.276 & 0.329 & 0.157 & 0.101 & 0.357
& 0.239 & 0.825
& 0.397 & 0.107 & 0.607
& 0.628 & 0.466 \\
\addlinespace[1pt]
PosterOmni~\citep{chen2026posteromni}
& 0.582 & 0.625 & 0.469 & 0.638 & 0.612
& 0.675 & 0.709 & 0.556 & 0.196 & 0.607
& 0.063 & 0.222
& 0.194 & 0.102 & 0.455
& 0.463 & 0.447 \\
\addlinespace[1pt]
CreatiDesign~\citep{zhang2026creatidesign}
& 0.674 & 0.731 & 0.585 & 0.750 & 0.725
& 0.210 & 0.273 & 0.148 & 0.083 & 0.345
& 0.080 & 0.267
& 0.515 & 0.329 & 0.670
& 0.561 & 0.426 \\
\midrule
\multicolumn{18}{l}{\textbf{\textit{Our Models}}} \\
\addlinespace[1pt]
\shortstack[l]{Compo-LongCat}
& \cellcolor{bestcolor}0.735 & \cellcolor{bestcolor}0.776 & \cellcolor{secondcolor}0.642 & \cellcolor{bestcolor}0.793 & \cellcolor{bestcolor}0.781
& \cellcolor{secondcolor}0.795 & \cellcolor{secondcolor}0.891 & \cellcolor{secondcolor}0.757 & \cellcolor{secondcolor}0.261 & \cellcolor{secondcolor}0.904
& \cellcolor{secondcolor}0.259 & \cellcolor{secondcolor}0.828
& \cellcolor{bestcolor}0.970 & \cellcolor{bestcolor}0.940 & \cellcolor{bestcolor}0.944
& \cellcolor{secondcolor}0.885 & \cellcolor{bestcolor}0.752 \\
\addlinespace[1pt]
\shortstack[l]{Compo-Qwen}
& \cellcolor{secondcolor}0.734 & \cellcolor{secondcolor}0.770 & \cellcolor{bestcolor}0.649 & \cellcolor{secondcolor}0.782 & \cellcolor{secondcolor}0.773
& \cellcolor{bestcolor}0.810 & \cellcolor{bestcolor}0.910 & \cellcolor{bestcolor}0.781 & \cellcolor{bestcolor}0.263 & \cellcolor{bestcolor}0.932
& \cellcolor{bestcolor}0.262 & \cellcolor{bestcolor}0.850
& \cellcolor{secondcolor}0.940 & \cellcolor{secondcolor}0.868 & \cellcolor{secondcolor}0.934
& \cellcolor{bestcolor}0.886 & \cellcolor{secondcolor}0.751 \\
\bottomrule
\end{tabular}%
}
\vspace{-3mm}
\endgroup
\end{table}

\textbf{Benchmark Data Creation.} 
We follow the training data generation pipeline to construct a curated benchmark of 500 samples with no overlap with the training set. Each sample is paired with structured annotations, enabling evaluation of both general text-and-image-to-image models and dedicated poster generation methods with different input formats. Details are included in Appendix~\ref{sec:benchmark-details}.

\textbf{Evaluation Metrics.} We evaluate each binding type using metrics tailored to its preservation requirements under a unified region matching protocol. For Semantic, Identity, and Pixel Binding, we use SAM-3~\citep{carion2026sam3} to segment generated instances, whereas for Text Binding, we use Chandra-OCR-2~\citep{chandra2} to detect rendered text and its spatial regions. Each detected instance or text region is matched to the ground-truth region with the highest IoU. A match is considered unsuccessful if its IoU falls below a predefined threshold, in which case all corresponding metrics are set to zero. For successfully matched regions, Semantic Binding is evaluated using CLIP-T~\citep{radford2021clip} to measure semantic alignment. Identity Binding is evaluated using CLIP-I~\citep{radford2021clip} and DINO-I~\citep{oquab2023dinov2} to measure identity similarity, together with CLIP-T to assess consistency with the corresponding Text Specifications. As Pixel Binding imposes stricter preservation requirements, allowing only variations in style and lighting while requiring shape and spatial location to remain unchanged, we additionally compute SSIM~\citep{wang2004ssim} alongside CLIP-I and DINO-I. Text Binding is evaluated using normalized edit distance (NED; i.e., one minus the edit distance)~\citep{gao2025postermaker} and sentence accuracy~\citep{zhang2026creatidesign}. All metrics are first computed for individual instances or text regions and then aggregated at the image level. We additionally report M-DINO~\citep{zhang2026creatidesign} for instance preservation, which penalizes poorly preserved instances more strongly and thus provides a stricter assessment than arithmetic averaging. The overall IoU and overall average score (Avg. Score) are computed as the means over all evaluated regions across binding types and all metrics, respectively.

\textbf{Implementation Details.} We initialize Compo with LongCat-Image-Edit~\citep{team2025longcatimage} and Qwen-Image-Edit-2511~\citep{wu2025qwenimage}, respectively. The resulting models are referred to as \textit{Compo-LongCat} and \textit{Compo-Qwen}. We use DiffusionNFT~\citep{zheng2026diffusionnft} as our RL post-training algorithm. More implementation details are provided in Appendices~\ref{sec:supervised-training-details} and~\ref{sec:RL-details}.

\subsection{Main Results}

\textbf{Quantitative Results.}
We compare our model against general text-and-image-to-image models and dedicated poster generation methods in Table~\ref{tab:benchmark}. For general text-and-image-to-image models, our benchmark is directly compatible with their standard input interfaces, using the Spatial Canvas, Text Specifications, and supplementary task instructions for accurate interpretation of input conditions. For dedicated poster generation methods, we convert the structured annotations into their respective input formats while preserving as much information as supported by their original protocols, enabling fair comparison without modifying their native capabilities. As shown in Table~\ref{tab:benchmark}, both Compo-LongCat and Compo-Qwen outperform all baselines on nearly all metrics, yielding consistent overall gains. Comparisons before and after RL post-training are provided in Appendix~\ref{sec:more-metric}.

\begin{figure}[!t]
\begin{center}
     \includegraphics[width=\textwidth]{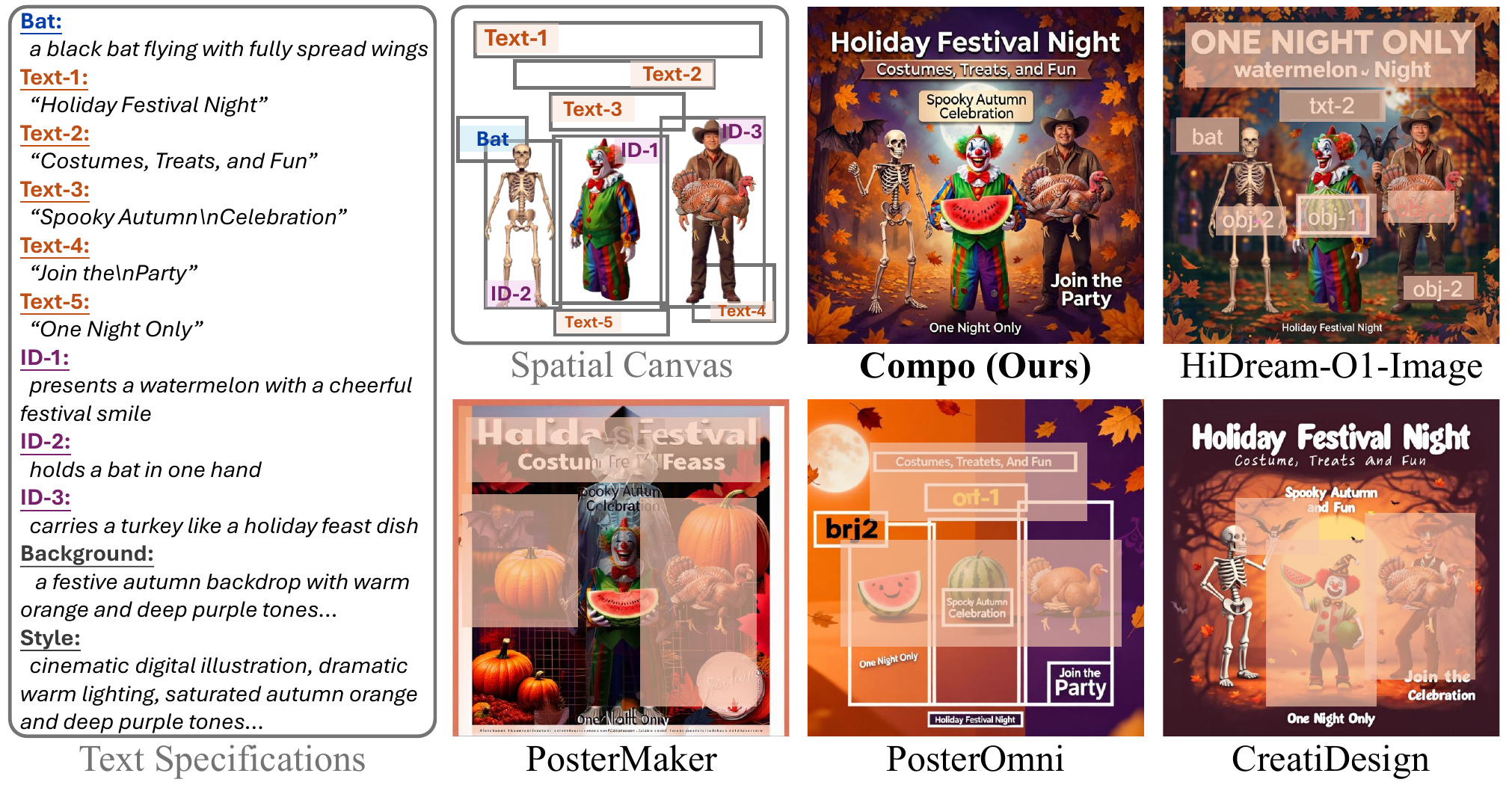}
\end{center}
\vspace{-5mm}
\caption{\textbf{Visual comparison with representative baselines.} We compare Compo with dedicated poster generation methods (PosterMaker~\citep{gao2025postermaker}, PosterOmni~\citep{chen2026posteromni}, and CreatiDesign~\citep{zhang2026creatidesign}) and the best-performing general text-and-image-to-image baseline in Table~\ref{tab:benchmark}, HiDream-O1-Image~\citep{cai2026hidream}. Transparent pink boxes highlight noticeable errors. Additional visualizations are provided in Appendix~\ref{sec:more-vis}.}
\vspace{-3mm}
\label{fig:qualitative_results}
\end{figure}

\textbf{Qualitative Results.} Figure~\ref{fig:qualitative_results} qualitatively compares Compo with four representative baselines. Additional comparisons and visualizations are illustrated in Appendix~\ref{sec:more-vis}.

\definecolor{bestcolor}{HTML}{FCE4D6}
\definecolor{secondcolor}{HTML}{D9EAF7}
\begin{table}[t]
\caption{\textbf{Effects of different RL reward components.}
We ablate the text, instance similarity (``Ins. Sim.''), and layout rewards individually and jointly. The \orangebox{best} and \bluebox{second-best} results are highlighted in orange and blue, respectively. ``NED'': normalized edit distance. ``Sen. Acc.'': sentence accuracy.}
\label{tab:ablation-reward-types}
\centering
\begingroup
\setlength{\tabcolsep}{2.3pt}
\renewcommand{\arraystretch}{1.12}
\resizebox{\linewidth}{!}{%
\begin{tabular}{@{}l*{17}{c}@{}}
\toprule
\multirow{2}{*}{\bfseries \shortstack[l]{Reward Setting}} & \multicolumn{5}{c}{\bfseries Pixel Binding}
& \multicolumn{5}{c}{\bfseries Identity Binding}
& \multicolumn{2}{c}{\bfseries Semantic Binding}
& \multicolumn{3}{c}{\bfseries Text Binding}
& \multicolumn{2}{c}{\bfseries Overall} \\
\cmidrule(lr){2-6}\cmidrule(lr){7-11}\cmidrule(lr){12-13}\cmidrule(lr){14-16}\cmidrule(lr){17-18}
& DINO-I & CLIP-I & M-DINO & SSIM & IoU
& DINO-I & CLIP-I & M-DINO & CLIP-T & IoU
& CLIP-T & IoU
& NED & \shortstack{Sen. Acc.} & IoU
& IoU & \shortstack{Avg. Score} \\
\midrule

Full Reward (Default)
& \cellcolor{secondcolor}0.735 & \cellcolor{secondcolor}0.776 & 0.642 & \cellcolor{secondcolor}0.793 & \cellcolor{secondcolor}0.781
& 0.795 & 0.891 & 0.757 & \cellcolor{secondcolor}0.261 & \cellcolor{secondcolor}0.904
& 0.259 & 0.828
& \cellcolor{bestcolor}0.970 & \cellcolor{bestcolor}0.940 & \cellcolor{bestcolor}0.944
& \cellcolor{bestcolor}0.885 & \cellcolor{bestcolor}0.752 \\

\addlinespace[1pt]
w/o Text Reward
& \cellcolor{bestcolor}0.752 & \cellcolor{bestcolor}0.789 & \cellcolor{bestcolor}0.666 & \cellcolor{bestcolor}0.801 & \cellcolor{bestcolor}0.790
& \cellcolor{bestcolor}0.811 & \cellcolor{bestcolor}0.901 & \cellcolor{bestcolor}0.776 & \cellcolor{bestcolor}0.262 & \cellcolor{bestcolor}0.928
& \cellcolor{bestcolor}0.275 & \cellcolor{bestcolor}0.884
& 0.878 & 0.778 & 0.895
& 0.876 & 0.746 \\

\addlinespace[1pt]
w/o Ins. Sim. Reward
& 0.725 & 0.769 & 0.639 & 0.788 & 0.776
& 0.782 & 0.879 & 0.740 & 0.255 & 0.877
& 0.239 & 0.753
& 0.958 & 0.924 & \cellcolor{secondcolor}0.942
& 0.866 & 0.736 \\
\addlinespace[1pt]
w/o Layout Reward
& 0.732 & 0.771 & \cellcolor{secondcolor}0.644 & 0.787 & 0.779
& \cellcolor{secondcolor}0.799 & \cellcolor{secondcolor}0.892 & \cellcolor{secondcolor}0.760 & 0.260 & 0.902
& \cellcolor{secondcolor}0.263 & \cellcolor{secondcolor}0.830
& \cellcolor{secondcolor}0.968 & \cellcolor{secondcolor}0.930 & 0.933
& \cellcolor{secondcolor}0.878 & \cellcolor{secondcolor}0.750 \\
\addlinespace[1pt]
w/o All Rewards
& 0.716 & 0.763 & 0.619 & 0.785 & 0.775
& 0.780 & 0.878 & 0.733 & 0.257 & 0.893
& 0.226 & 0.720
& 0.924 & 0.886 & 0.916
& 0.849 & 0.725 \\
\bottomrule
\end{tabular}%
}
\vspace{-3mm}
\endgroup
\end{table}

\subsection{Ablation Study}
We conduct all ablation studies using Compo-LongCat for training and evaluation efficiency.

\textbf{Ablation on RL Reward Components.}
As introduced in Section~\ref{sec:architecture}, our RL objective consists of three complementary rewards: text reward, instance similarity reward, and layout reward. We ablate each component individually and also remove all rewards to study their contributions. As shown in Table~\ref{tab:ablation-reward-types}, removing the text reward substantially degrades text binding, while removing the instance similarity reward mainly affects pixel- and identity-level consistency. The layout reward is important for spatial alignment, as its removal leads to a drop in IoU-related metrics. Removing all rewards results in the lowest overall performance, confirming that the three reward components provide complementary supervision during RL post-training.

\definecolor{bestcolor}{HTML}{FCE4D6}
\definecolor{secondcolor}{HTML}{D9EAF7}
\begin{wraptable}{r}{0.5\textwidth}
\centering
\vspace{-5mm}
\caption{\textbf{Effects of different sentence accuracy reward weights.} Smaller weights tolerate minor text rendering errors, while larger weights over-rely on the sparse sentence accuracy during the early stage of RL post-training. The \orangebox{best} and \bluebox{second-best} results are highlighted in orange and blue, respectively. ``NED'': normalized edit distance. ``Sen. Acc.'': sentence accuracy.}
\label{tab:ablation-text-reward}
\small
\begingroup
\setlength{\tabcolsep}{4pt}
\renewcommand{\arraystretch}{1.08}
\begin{tabular}{@{}l*{3}{c}@{}}
\toprule
\multirow{2}{*}{\bfseries \shortstack[l]{Sen. Acc. Weight}} & 
\multicolumn{3}{c}{\bfseries Text Binding} \\
\cmidrule(lr){2-4}
& NED & \shortstack{Sen. Acc.} & IoU \\
\midrule
0.75 (Default)
& \cellcolor{secondcolor}0.970
& \cellcolor{bestcolor}0.940
& \cellcolor{secondcolor}0.944 \\
\addlinespace[1pt]
0.0
& \cellcolor{bestcolor}0.971 & 0.916 & \cellcolor{bestcolor}0.945 \\
\addlinespace[1pt]
0.25
& 0.966 & 0.920 & \cellcolor{secondcolor}0.944 \\
\addlinespace[1pt]
0.5
& 0.966 & 0.928 & \cellcolor{bestcolor}0.945 \\
\addlinespace[1pt]
1.0
& 0.969 & \cellcolor{secondcolor}0.938 & 0.938 \\
\bottomrule
\end{tabular}
\vspace{-2mm}
\endgroup
\end{wraptable}

\textbf{Ablation on Text Reward Metrics.} We further ablate the detailed metrics of the text reward, as text rendering errors can substantially compromise the usability of a poster. Similar to evaluation metrics, our text reward combines NED and sentence accuracy. Both metrics are first computed for each bound text element and then aggregated at the image level. NED, a continuous score in $[0,1]$, is aggregated by arithmetic averaging. Sentence accuracy is binary and aggregated using a logical AND operation, assigning a score of one only when all bound text elements are rendered correctly. It therefore provides a stricter measure of poster-level text correctness. The combined text reward is defined as $ R_{\mathrm{text}} = w R_{\mathrm{Sen.Acc.}} + (1-w) R_{\mathrm{NED}}$, where $w$ controls the contribution of sentence accuracy. We evaluate the effects of different weights $w \in \{0, 0.25, 0.5, 0.75, 1\}$. As shown in Table~\ref{tab:ablation-text-reward}, $w=0.75$ achieves the highest sentence accuracy and the second-highest NED without compromising layout adherence. Smaller weights, such as $0$ and $0.25$, overemphasize NED, allowing minor rendering errors to receive relatively high rewards and weakening the supervision for fully correct poster text. Conversely, setting $w=1$ relies exclusively on sentence accuracy. During the early stage of reinforcement learning, rollout candidates generated with a guidance scale of $1.0$ rarely render all text elements correctly, causing sentence accuracy to be frequently zero and producing an overly sparse reward. Combining the two metrics therefore balances dense character-level feedback with strict poster-level correctness.

\definecolor{bestcolor}{HTML}{FCE4D6}
\definecolor{secondcolor}{HTML}{D9EAF7}
\begin{table}[h]
\vspace{-2mm}
\caption{\textbf{Effects of different LoRA target modules.} The \orangebox{best} and \bluebox{second-best} results are highlighted in orange and blue, respectively. ``NED'': normalized edit distance. ``Sen. Acc.'': sentence accuracy.}
\label{tab:ablation-lora-modules}
\centering
\begingroup
\setlength{\tabcolsep}{2.3pt}
\renewcommand{\arraystretch}{1.12}
\resizebox{\linewidth}{!}{%
\begin{tabular}{@{}l*{17}{c}@{}}
\toprule
\multirow{2}{*}{\bfseries \shortstack[l]{LoRA Target Modules}} & \multicolumn{5}{c}{\bfseries Pixel Binding}
& \multicolumn{5}{c}{\bfseries Identity Binding}
& \multicolumn{2}{c}{\bfseries Semantic Binding}
& \multicolumn{3}{c}{\bfseries Text Binding}
& \multicolumn{2}{c}{\bfseries Overall} \\
\cmidrule(lr){2-6}\cmidrule(lr){7-11}\cmidrule(lr){12-13}\cmidrule(lr){14-16}\cmidrule(lr){17-18}
& DINO-I & CLIP-I & M-DINO & SSIM & IoU
& DINO-I & CLIP-I & M-DINO & CLIP-T & IoU
& CLIP-T & IoU
& NED & \shortstack{Sen. Acc.} & IoU
& IoU & \shortstack{Avg. Score} \\
\midrule
Both Branches (Default)
& \cellcolor{secondcolor}0.716 & 0.763 & 0.619 & 0.785 & 0.775
& \cellcolor{bestcolor}0.780 & \cellcolor{bestcolor}0.878 & \cellcolor{secondcolor}0.733 & \cellcolor{bestcolor}0.257 & \cellcolor{bestcolor}0.893
& \cellcolor{secondcolor}0.226 & \cellcolor{secondcolor}0.720
& \cellcolor{secondcolor}0.924 & \cellcolor{secondcolor}0.886 & \cellcolor{bestcolor}0.916
& \cellcolor{secondcolor}0.849 & \cellcolor{bestcolor}0.725 \\
\addlinespace[1pt]
Image Branch Only
& 0.712 & \cellcolor{bestcolor}0.764 & 0.623 & \cellcolor{bestcolor}0.787 & \cellcolor{secondcolor}0.777
& 0.772 & 0.873 & 0.731 & \cellcolor{secondcolor}0.256 & 0.881
& \cellcolor{bestcolor}0.226 & 0.710
& 0.913 & 0.878 & 0.906
& 0.841 & 0.721 \\
\addlinespace[1pt]
Text Branch Only
& 0.714 & \cellcolor{secondcolor}0.764 & \cellcolor{bestcolor}0.626 & \cellcolor{secondcolor}0.786 & 0.775
& 0.770 & 0.868 & 0.722 & 0.253 & 0.882
& 0.224 & 0.711
& \cellcolor{bestcolor}0.926 & \cellcolor{bestcolor}0.891 & \cellcolor{secondcolor}0.916
& 0.846 & 0.722 \\
\bottomrule
\end{tabular}%
}
\vspace{-1mm}
\endgroup
\end{table}

\textbf{Ablation on LoRA Target Modules.}
Our default setting applies LoRA to both the text and image branches of the pretrained MMDiT, adapting its original image-editing capability to poster generation through the Spatial Canvas Interface. As shown in Table~\ref{tab:ablation-lora-modules}, fine-tuning both branches achieves the best overall performance. The two branches play complementary roles in our framework: the text branch needs to adapt to the structured VL tokens that encode the Text Specifications and Spatial Canvas information, while the image branch must learn to jointly interpret the noisy target tokens and spatially organized canvas tokens for composition-aware generation. Fine-tuning only one branch leaves the other insufficiently adapted to the new conditioning interface, resulting in consistently weaker binding performance.

\section{Conclusion}
\vspace{-1mm}
In this work, we introduce Compo, a poster generation model built around a Spatial Canvas Interface that enables users to specify generation intent through explicit spatial composition rather than text prompting alone. The interface unifies semantic, identity, text, and pixel bindings with accompanying Text Specifications, providing fine-grained control over poster content, layout, and appearance. Compo is efficiently adapted from pretrained image editing models using automatically generated structured supervision and binding-aware reinforcement learning, and supports both direct composition and agentic generation from high-level user requests. Experiments show that Compo achieves stronger compositional controllability than both general-purpose image generation models and dedicated poster generation systems while maintaining high visual quality.

\newpage

\subsection*{AI use statement}
In this work, we used large language models to assist data generation and annotation, including text asset preparation, prompt writing, instance captioning, and agentic planning, and image generation models to create target images for the dataset and benchmark. Additionally, we used large language models to improve the clarity and readability of the manuscript. We performed a manual human review of the LLM-generated text asset repositories to ensure they contain no toxic, harmful, or inappropriate content. Additionally, while large language models were utilized for language refinement during manuscript preparation, all contents were thoroughly reviewed and verified by the authors to guarantee they faithfully represent our original research ideas and experimental results. We take responsibility for the final content of this work, including text, claims, or artifacts produced with the aid of generative AI.

\subsection*{Ethics statement}
Our data consists of images sourced from publicly available image generation models. All assets used during experiments and manuscript writing are either obtained from publicly accessible Internet sources or created by generative models. We encourage responsible use of our models and data that respects user privacy and the intellectual property rights of image owners, particularly in commercial applications.

\subsection*{Reproducibility statement}
To facilitate reproducibility, Sections~\ref{sec:data},~\ref{sec:architecture},~\ref{sec:inference}, and~\ref{sec:experiments} describe the dataset construction, model training, inference pipeline, and evaluation settings, respectively. Appendices~\ref{sec:dataset-details},~\ref{sec:benchmark-details}, and~\ref{sec:imp-details} provide additional details on dataset and benchmark construction, supervised training, and reinforcement learning.

\bibliography{iclr2027_conference}
\bibliographystyle{iclr2027_conference}

\appendix
\newpage

{\Large Appendix}

\section{Dataset and Benchmark Details}

\subsection{Compo-200K Dataset}
\label{sec:dataset-details}

\textbf{Text Asset Collection.}
Leveraging advanced large language models (LLMs), including \textit{GPT-5.5}~\citep{gpt55}, \textit{Claude Opus 4.6}~\citep{claudeopus46}, and \textit{Gemini 3.5 Flash}~\citep{gemini35flash}, we curate two distinct text asset repositories to ensure broad semantic and visual coverage. Specifically, the content repository contains roughly 1,000 instances spanning 12 poster themes such as product advertising, public service, and entertainment, while the style repository consists of around 250 fine-grained visual style descriptions.

\textbf{LLM Planning.}
To construct each training sample, we first sample a poster theme, $3\text{--}8$ associated instances, and a visual style from our repositories. Guided by system prompts in Section~\ref{sec:prompt-llm-plan}, GPT-5.5 then plans additional design elements required for poster generation as described in Section~\ref{sec:data}, and produces a structured JSON specification that integrates the sampled assets with the newly planned elements. Ultimately, this procedure yields approximately 200K structured plans, which serve as inputs for subsequent prompt rewriting to facilitate downstream poster data generation.

\textbf{LLM Rewriting.}
Guided by system prompts in Section~\ref{sec:prompt-llm-rewrite}, GPT-5.5 converts each structured JSON specification into a natural-language poster generation prompt, faithfully preserving both sampled assets and planned design attributes.

\textbf{Text-to-Poster Generation.} 
To synthesize poster images, we employ three text-to-image models, including \textit{Qwen-Image-2512}~\citep{wu2025qwenimage}, \textit{Z-Image}~\citep{cai2025z}, and \textit{ERNIE-Image}~\citep{liu2026ernie}, as generation backends, capitalizing on their superior instruction following, text rendering, and graphic design capabilities. For each prompt, we randomly select one model to render a poster at $1024^2$ resolution, thereby enhancing dataset diversity and mitigating model-specific biases. The data generation pipeline is conducted on approximately $32\times$ NVIDIA A100 80GB GPUs.

\textbf{Instance Segmentation, Relighting, and Reposing.} 
Leveraging semantic labels from the structured plans, we perform text-guided segmentation via \textit{SAM-3}~\citep{carion2026sam3} to extract visual instances. To standardize these extracted assets, we employ \textit{FLUX.2-klein-9B}~\citep{flux-2-2025} to normalize their pose or illumination: identity instances are reposed into predefined canonical postures (e.g., front-facing, arm-folded, three-quarter standing, and half-length portraits), whereas object instances undergo flat relighting to eliminate specular highlights and directional shadows.

\textbf{Text Recognition.} 
To extract visual text alongside its spatial layout, we adopt \textit{Chandra-OCR-2}~\citep{chandra2}, a specialized vision-language model built upon \textit{Qwen3.5}~\citep{qwen3.5}, for joint text recognition and region localization. To enforce strict alignment with the rendered poster and mitigate potential generation discrepancies, we adopt the OCR-recognized text, rather than the initial specification in the structured plan, as the final ground-truth annotation.

\textbf{Spatial Canvas Rendering and Text Specification Formulation.} Given the instance segmentation and text recognition results, we construct a black Spatial Canvas matching the resolution of the target poster. Visual instance locations are derived from segmentation masks, while text regions are extracted directly from recognition results. We populate the canvas with four binding types: for \textit{Text Binding}, each region is framed by a white bounding box labeled with an indexed identifier, ``txt-k''; for \textit{Identity Binding}, we place the reposed identity reference within its bounding box, tagged with ``obj-k''. Object instances are randomly assigned to either \textit{Pixel Binding} (where the reference image is directly pasted without extra annotation) or \textit{Semantic Binding} (where only a bounding box containing the class name, e.g., ``lemon'', is rendered). All identifiers, bounding boxes, and reference images are rendered directly on the Spatial Canvas, while their detailed descriptions are systematically documented in the Text Specifications.

\subsection{Benchmark}
\label{sec:benchmark-details}
\textbf{Benchmark Data Creation.} To construct each evaluation sample, we randomly sample a theme, $4\text{--}9$ associated instances, and a visual style using random seeds different from those used for training data generation. The number of instances is intentionally larger than the $3\text{--}8$ instances used during training data generation, making the benchmark more challenging. Following the training data generation pipeline while ensuring no overlap with the training set, we curate a benchmark of 500 samples. Each sample is paired with a Spatial Canvas and detailed structured annotations.

\textbf{Evaluation Metrics.} During evaluation, each model is provided with inputs in its expected format. For general text-and-image-to-image models, the inputs are directly constructed from the Spatial Canvas, Text Specifications, and supplementary task instructions for accurate interpretation of the input conditions. For dedicated poster generation methods, the structured benchmark annotations are converted into the respective input formats prescribed by each method, while preserving as much of the original information as supported by their native input protocols. The benchmark metrics are then computed fully automatically from the generated poster images and the corresponding pre-constructed structured annotations, including instance images, rendered text, and layout bounding boxes. This evaluation protocol is fully model-agnostic and requires neither model-specific input designs nor access to model-specific internal representations.

\section{Implementation Details}
\label{sec:imp-details}

\subsection{Supervised Training}
\label{sec:supervised-training-details}
We initialize our models from LongCat-Image-Edit~\citep{team2025longcatimage} and Qwen-Image-Edit-2511~\citep{wu2025qwenimage}, and fine-tune them using LoRA~\citep{hu2021lora} with a rank of 128. We employ aspect-ratio bucketing with a target resolution of approximately $1024^2$ pixels. For LongCat-Image-Edit, we use $8\times$ NVIDIA A100 80GB GPUs with a global batch size of 64, while Qwen-Image-Edit-2511 is trained on $16\times$ NVIDIA A100 80GB GPUs with a global batch size of 32. For both models, we use a constant learning rate of 1e-4 with 500 warmup steps.

\subsection{Reinforcement Learning}
\label{sec:RL-details}

Both models are initialized from their respective supervised training checkpoints, using a newly introduced LoRA module with a reduced rank of 64. During rollout generation, each input sample yields 10 candidates using 10 denoising steps at a guidance scale of 1.0. Specifically, LongCat-Image-Edit operates at a resolution of $1024^2$ pixels, whereas Qwen-Image-Edit-2511 is configured at a lower resolution of $512^2$ to alleviate GPU memory overhead. Both models are trained on $16\times$ NVIDIA A100 80GB GPUs.

For each rollout candidate, we evaluate the generated image using the same region-matching protocol as our benchmark. We employ SAM-3~\citep{carion2026sam3} to segment visual instances and Chandra-OCR-2~\citep{chandra2} to detect rendered text and its corresponding spatial regions. Each detected visual or text region is matched to its ground-truth counterpart based on maximum IoU. If the resulting IoU falls below a predefined threshold, the match is considered invalid, and all associated rewards are set to zero. For valid matches, we compute element-level rewards according to their binding types. For Text Binding, the text reward is based on sentence accuracy and normalized edit distance. For Semantic Binding, we use CLIP-T~\citep{radford2021clip}; for Identity Binding, we use CLIP-I~\citep{radford2021clip}, DINO-I~\citep{oquab2023dinov2}, and CLIP-T; and for Pixel Binding, we use CLIP-I and DINO-I. The layout reward is defined as the IoU between each matched region and its ground-truth location and is computed for both visual and textual elements.

We aggregate the element-level rewards into an image-level reward for each rollout candidate. Sentence accuracy is aggregated using a logical AND operation, yielding one only when all bound text elements are rendered correctly, while all other reward metrics are aggregated by arithmetic averaging over the corresponding elements. We then combine the image-level text, similarity, and layout rewards to optimize the trainable LoRA modules using the DiffusionNFT~\citep{zheng2026diffusionnft} objective. Specifically, the overall reward is defined as
\begin{equation}
R = \lambda_{\mathrm{Text}} R_{\mathrm{Text}} + \lambda_{\mathrm{Ins.Sim.}} R_{\mathrm{Ins.Sim.}} + \lambda_{\mathrm{Layout}} R_{\mathrm{Layout}},
\end{equation}
where $\lambda_{\mathrm{Text}}$, $\lambda_{\mathrm{Ins.Sim.}}$, and $\lambda_{\mathrm{Layout}}$ control the contributions of text rendering, instance similarity, and layout adherence, respectively. We set $\lambda_{\mathrm{Text}}=0.4$, $\lambda_{\mathrm{Ins.Sim.}}=0.4$, and $\lambda_{\mathrm{Layout}}=0.2$ throughout RL post-training.

\subsection{Agentic Mode}
\label{sec:agentic-mode}

Our Agentic Mode adopts a workflow-based pipeline with verification of image search results. We use \textit{GPT-5.6}~\citep{gpt56} as both the planner and verifier. Using the prompts detailed in Section~\ref{sec:prompt-agent}, GPT-5.6 extracts the visual instances, determines their binding types, and infers their spatial relationships. For instances requiring image search, it further verifies the relevance and correctness of the retrieved images.

\section{More Results}

\subsection{Quantitative Results}
\label{sec:more-metric}
We additionally compare our Compo models before and after RL post-training. As shown in Table~\ref{tab:sft-rl-compare}, \textit{Compo-LongCat-Base} and \textit{Compo-Qwen-Base} denote the models after supervised training, whereas \textit{Compo-LongCat} and \textit{Compo-Qwen} denote their corresponding models after RL post-training. Our RL objective incorporates three complementary reward components, providing diverse supervision signals during post-training. As a result, RL post-training yields consistent improvements across most evaluation metrics for both model variants.

\definecolor{bestcolor}{HTML}{FCE4D6}
\definecolor{secondcolor}{HTML}{D9EAF7}
\begin{table}[h]
\caption{We compare our Compo models before and after RL post-training across pixel, identity, semantic, and text binding metrics. Higher values indicate better performance. The \orangebox{best} and \bluebox{second-best} results are highlighted in orange and blue, respectively. ``NED'': normalized edit distance. ``Sen. Acc.'': sentence accuracy.}
\label{tab:sft-rl-compare}
\centering
\begingroup
\setlength{\tabcolsep}{2.3pt}
\renewcommand{\arraystretch}{1.12}
\resizebox{\linewidth}{!}{%
\begin{tabular}{@{}l*{17}{c}@{}}
\toprule
\multirow{2}{*}{\bfseries \shortstack[l]{Models}} & \multicolumn{5}{c}{\bfseries Pixel Binding}
& \multicolumn{5}{c}{\bfseries Identity Binding}
& \multicolumn{2}{c}{\bfseries Semantic Binding}
& \multicolumn{3}{c}{\bfseries Text Binding}
& \multicolumn{2}{c}{\bfseries Overall} \\
\cmidrule(lr){2-6}\cmidrule(lr){7-11}\cmidrule(lr){12-13}\cmidrule(lr){14-16}\cmidrule(lr){17-18}
& DINO-I & CLIP-I & M-DINO & SSIM & IoU
& DINO-I & CLIP-I & M-DINO & CLIP-T & IoU
& CLIP-T & IoU
& NED & \shortstack{Sen. Acc.} & IoU
& IoU & \shortstack{Avg. Score} \\
\midrule
\addlinespace[1pt]
\shortstack[l]{Compo-LongCat-Base}
& 0.716 & 0.763 & 0.619 & \cellcolor{secondcolor}0.785 & \cellcolor{secondcolor}0.775
& 0.780 & 0.878 & 0.733 & 0.257 & 0.893
& 0.226 & 0.720
& 0.924 & \cellcolor{secondcolor}0.886 & 0.916
& 0.849 & 0.725 \\
\addlinespace[1pt]
Compo-Qwen-Base
& 0.720 & 0.765 & 0.631 & 0.785 & 0.772
& 0.791 & 0.891 & \cellcolor{secondcolor}0.758 & 0.259 & \cellcolor{secondcolor}0.918
& 0.217 & 0.716
& 0.907 & 0.858 & 0.906
& 0.845 & 0.726 \\
\addlinespace[1pt]
\shortstack[l]{Compo-LongCat}
& \cellcolor{bestcolor}0.735 & \cellcolor{bestcolor}0.776 & \cellcolor{secondcolor}0.642 & \cellcolor{bestcolor}0.793 & \cellcolor{bestcolor}0.781
& \cellcolor{secondcolor}0.795 & \cellcolor{secondcolor}0.891 & 0.757 & \cellcolor{secondcolor}0.261 & 0.904
& \cellcolor{secondcolor}0.259 & \cellcolor{secondcolor}0.828
& \cellcolor{bestcolor}0.970 & \cellcolor{bestcolor}0.940 & \cellcolor{bestcolor}0.944
& \cellcolor{secondcolor}0.885 & \cellcolor{bestcolor}0.752 \\
\addlinespace[1pt]
\shortstack[l]{Compo-Qwen}
& \cellcolor{secondcolor}0.734 & \cellcolor{secondcolor}0.770 & \cellcolor{bestcolor}0.649 & 0.782 & 0.773
& \cellcolor{bestcolor}0.810 & \cellcolor{bestcolor}0.910 & \cellcolor{bestcolor}0.781 & \cellcolor{bestcolor}0.263 & \cellcolor{bestcolor}0.932
& \cellcolor{bestcolor}0.262 & \cellcolor{bestcolor}0.850
& \cellcolor{secondcolor}0.940 & 0.868 & \cellcolor{secondcolor}0.934
& \cellcolor{bestcolor}0.886 & \cellcolor{secondcolor}0.751 \\
\bottomrule
\end{tabular}%
}
\vspace{-3mm}
\endgroup
\end{table}

\subsection{Qualitative Results}
\label{sec:more-vis}
\begin{figure}[!t]
\begin{center}
     \includegraphics[width=\textwidth]{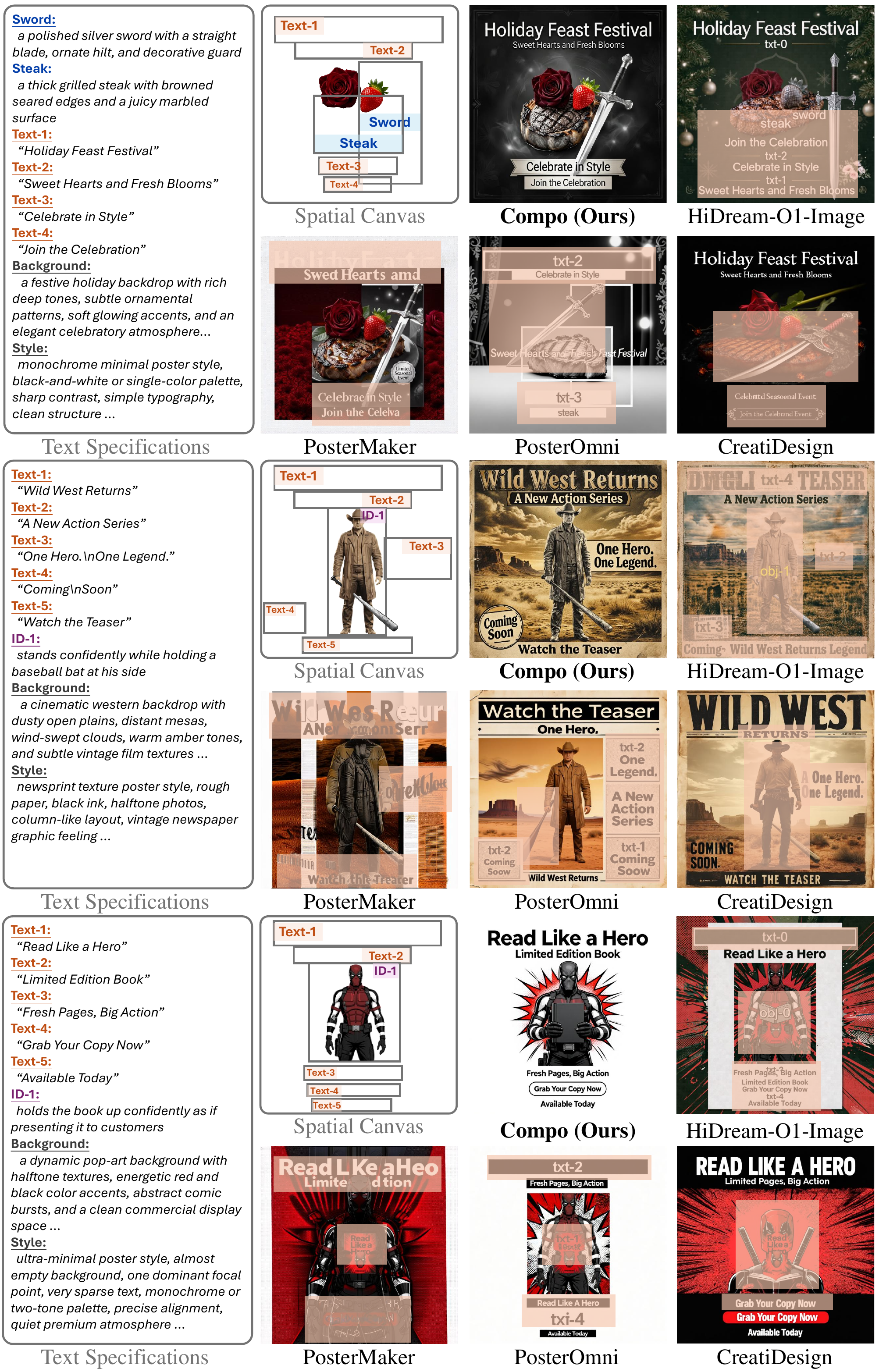}
\end{center}
\vspace{-5mm}
\caption{\textbf{Additional visual comparison with representative baselines (Part 1).} We compare Compo with dedicated poster generation methods (PosterMaker~\citep{gao2025postermaker}, PosterOmni~\citep{chen2026posteromni}, and CreatiDesign~\citep{zhang2026creatidesign}) and the best-performing general text-and-image-to-image baseline in Table~\ref{tab:benchmark}, HiDream-O1-Image~\citep{cai2026hidream}. Transparent pink boxes highlight noticeable errors.}
\vspace{-3mm}
\label{fig:qualitative_results_supp_0}
\end{figure}

\begin{figure}[!t]
\begin{center}
     \includegraphics[width=\textwidth]{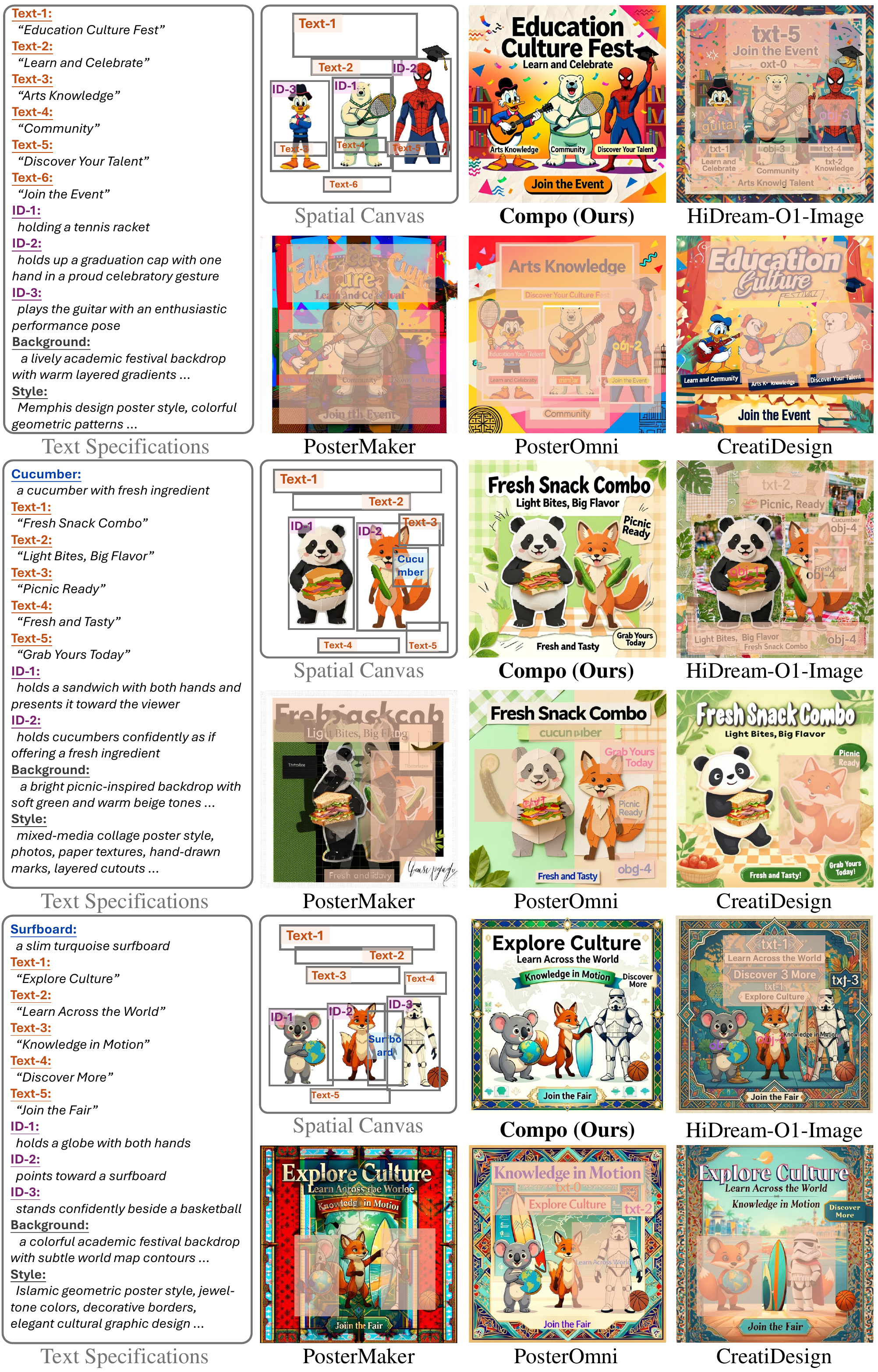}
\end{center}
\vspace{-5mm}
\caption{\textbf{Additional visual comparison with representative baselines (Part 2).} We compare Compo with dedicated poster generation methods (PosterMaker~\citep{gao2025postermaker}, PosterOmni~\citep{chen2026posteromni}, and CreatiDesign~\citep{zhang2026creatidesign}) and the best-performing general text-and-image-to-image baseline in Table~\ref{tab:benchmark}, HiDream-O1-Image~\citep{cai2026hidream}. Transparent pink boxes highlight noticeable errors.}
\vspace{-3mm}
\label{fig:qualitative_results_supp_1}
\end{figure}

\begin{figure}[!t]
\begin{center}
     \includegraphics[width=\textwidth]{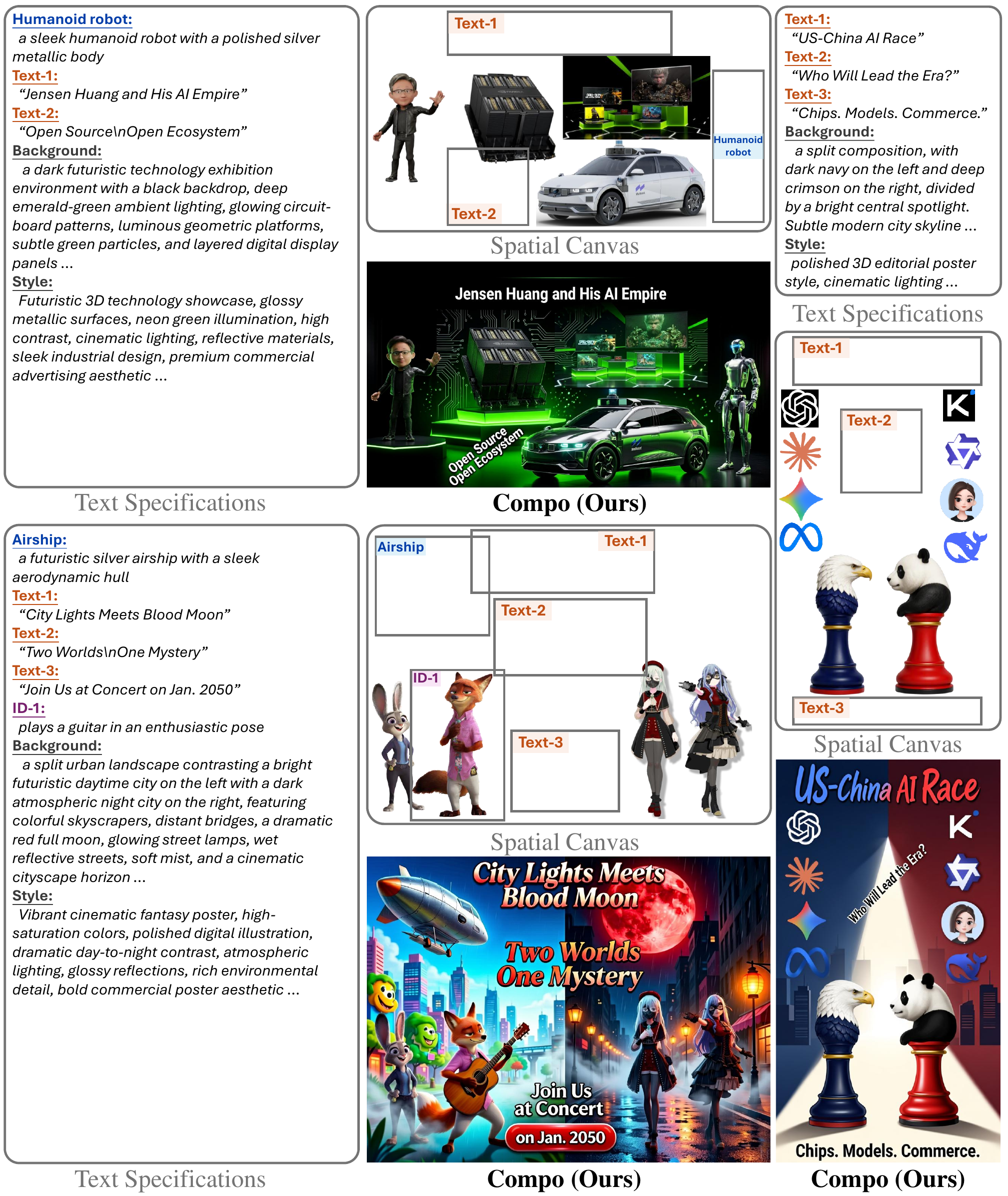}
\end{center}
\vspace{-5mm}
\caption{\textbf{Additional visualizations at various resolutions across diverse trending themes and visual elements.} Our Compo models can generate visually coherent and plausible poster designs based on diverse trending themes and visual elements that are unseen during training, demonstrating robust generalization and practical applicability.}
\vspace{-3mm}
\label{fig:qualitative_results_supp_2}
\end{figure}

We provide additional visual comparisons with representative baselines in Figures~\ref{fig:qualitative_results_supp_0} and~\ref{fig:qualitative_results_supp_1}. To further assess the generalization of our Compo models, Figure~\ref{fig:qualitative_results_supp_2} presents results at various resolutions across diverse trending themes and visual elements, including cutting-edge technology, Disney and Japanese animation, and AI industry news. All these themes and elements are unseen during training. Despite this distribution shift, our models consistently produce visually coherent and plausible poster designs, demonstrating robust generalization and practical applicability.

\clearpage

\section{Prompt Gallery}
\subsection{Prompt for LLM Planning in Dataset Generation}
\label{sec:prompt-llm-plan}

\begin{promptbox}{LLM Planning System Prompt (Identity-Object Interaction)}
You are a poster semantic planning agent.

Your task is to design a coherent poster concept based on the given input:
"Humanoids: [x1, x2, x3, ...], Things: [y1, y2, y3, ...], Theme: xxxx"

You must generate a structured semantic plan for a poster image. The plan will later be used by another model to write a full image-generation prompt.

Your output must be a valid JSON object only. Do not include markdown, comments, explanations, or any text outside the JSON.

Output schema:
{
  "scenario": string,
  "bg_prompt": string,
  "visual_text": [string, string, string, string, string, ...],
  "HOI": {
    "<humanoid_name>": {
      "action": string,
      "target_objects": [string, ...],
      "description": string
    }
  }
}

Planning requirements:

1. Scenario
- Create one specific poster scenario that matches the given Theme and all input objects.
- The scenario should describe the poster's concrete purpose, such as a product campaign, event promotion, public service message, entertainment poster, travel ad, education campaign, sports promotion, fashion campaign, or festival poster.
- The scenario must be coherent with the given Humanoids and Things.
- Do not make the scenario too generic. Avoid vague phrases like "a creative poster" or "an interesting advertisement".
- The scenario should be one concise sentence.

2. Background Prompt
- Generate one concise bg_prompt that describes the overall background of the poster.
- The bg_prompt should support the Theme and scenario through setting, atmosphere, color mood, graphic patterns, environmental context, depth, or decorative background elements.
- The bg_prompt must describe background only, not foreground subjects or products.
- The bg_prompt must not contain any existing foreground object from the input Humanoids or Things lists.
- Do not mention, repeat, or imply the exact names of any input Humanoids or Things in bg_prompt.
- Do not include human-object interactions, actions, product placement, poster text, slogans, or typography in bg_prompt.
- Keep bg_prompt visually useful, safe, concise, and suitable for image generation.

3. Visual Text
- Generate **3 to 5 short visual text phrases** that can appear on the poster.
- The visual text should match the Theme, scenario, and objects.
- Include a mix of poster text roles, such as:
  - main headline
  - subheadline
  - slogan
  - call-to-action
  - badge text
  - short feature phrase
  - event/promo phrase
  - date/location phrase
- Each text phrase should be short and readable.
- Prefer natural poster-style English.
- Avoid overly long sentences.
- Avoid uncommon symbols, complex punctuation, or hard-to-render typography.
- Do not include explanations of text roles; output only the text strings in the list.
- The text should be visually useful for a poster, not a paragraph.

4. HOI
- For each humanoid in Humanoids, create exactly one HOI entry.
- The HOI key must exactly match the humanoid name from the input.
- Each humanoid should have a clear action or pose.
- The action should support the Theme and scenario.
- If possible, create meaningful human-object interactions between humanoids and Things.
- A humanoid may interact with zero, one, or multiple Things.
- "target_objects" must only contain objects from the input Things list.
- If the humanoid does not interact with any Thing, set "target_objects": [].
- Do not invent target objects that are not in Things.
- Do not invent additional humanoids.

5. Object Usage
- Try to make every Thing visually relevant in the scenario.
- Important Things should be used as products, props, foreground decorations, or interaction targets.
- If there are more Things than can be naturally used in HOI, some Things may be implied by the scenario or visual text, but HOI target_objects must still only include actual interaction targets.

6. Action Design
- Prefer concrete, visually grounded actions, such as:
  holding, presenting, drinking, eating, pointing at, riding, wearing, carrying, playing, reading, using, waving, giving a thumbs-up, posing confidently, dancing, running, protecting, teaching, performing.
- Avoid abstract actions that are hard to visualize, such as:
  appreciating, enjoying, supporting, representing, symbolizing, promoting.
- Avoid unsafe, violent, sexual, hateful, or disturbing actions.
- Avoid actions that are physically impossible or visually confusing.
- Keep actions simple enough for image generation.

7. Style of Descriptions
- The "description" field should be a clear visual sentence.
- Mention the target objects naturally if target_objects is non-empty.
- Use present tense.
- Avoid excessive detail.
- Avoid camera, lighting, rendering style, or layout descriptions in HOI; these belong to the later full prompt stage.
- Keep each description concise but specific.

8. Consistency
- The JSON must be valid and parseable.
- Do not add trailing commas.
- Do not use duplicated keys.
- Do not omit bg_prompt.
- Do not omit any humanoid from HOI.
- Do not include any input Humanoid or Thing name inside bg_prompt.
- Preserve exact spelling and capitalization of input object names.
- The output should be semantically coherent and suitable for generating a poster image.

Example input:
Humanoids: ["Iron Man", "panda", "polar bear"]
Things: ["coca-cola", "orange", "watermelon"]
Theme: Product_Advertising

Example output:
{
  "scenario": "A cheerful online shopping product advertisement promoting refreshing drinks and summer fruits with heroic mascot characters.",
  "bg_prompt": "a bright summer shopping backdrop with clean geometric shapes, warm sunshine gradients, fresh color accents, and spacious commercial display atmosphere",
  "visual_text": [
    "Hero Fresh Deals",
    "Cool Drinks, Fresh Fruits",
    "Refresh Your Cart",
    "Limited Summer Offer",
    "Shop Now"
  ],
  "HOI": {
    "Iron Man": {
      "action": "holding",
      "target_objects": ["coca-cola", "orange"],
      "description": "holds a bottle of coca-cola in his left hand and an orange in his right hand"
    },
    "panda": {
      "action": "thumbs up",
      "target_objects": [],
      "description": "gives a cheerful thumbs-up gesture toward the viewer"
    },
    "polar bear": {
      "action": "eating",
      "target_objects": ["watermelon"],
      "description": "eats a slice of watermelon happily"
    }
  }
}
\end{promptbox}

\begin{promptbox}{LLM Planning System Prompt (Object Only)}
You are a poster semantic planning agent.

Your task is to design a coherent object-only poster concept based on the given input:
"Things: [y1, y2, y3, ...], Theme: xxxx"

You must generate a structured semantic plan for a poster image. The plan will later be used by another model to write a full image-generation prompt.

Your output must be a valid JSON object only. Do not include markdown, comments, explanations, or any text outside the JSON.

Output schema:
{
  "scenario": string,
  "bg_prompt": string,
  "visual_text": [string, string, string, string, string, ...],
  "obj_only": {
    "<thing_name>": {
      "description": string
    }
  }
}

Planning requirements:

1. Scenario
- Create one specific poster scenario that matches the given Theme and all input Things.
- The scenario should describe the poster's concrete purpose, such as a product campaign, event promotion, public service message, entertainment poster, travel ad, education campaign, sports promotion, fashion campaign, or festival poster.
- The scenario must be coherent with the given Things.
- Do not make the scenario too generic. Avoid vague phrases like "a creative poster" or "an interesting advertisement".
- The scenario should be one concise sentence.

2. Background Prompt
- Generate one concise bg_prompt that describes the overall background of the poster.
- The bg_prompt should support the Theme and scenario through setting, atmosphere, color mood, graphic patterns, environmental context, depth, or decorative background elements.
- The bg_prompt must describe background only, not foreground products, props, or display items.
- The bg_prompt must not contain any existing foreground object from the input Things list.
- Do not mention, repeat, or imply the exact names of any input Things in bg_prompt.
- Do not include product placement, object display, poster text, slogans, typography, or object-specific attributes in bg_prompt.
- Keep bg_prompt visually useful, safe, concise, and suitable for image generation.

3. Visual Text
- Generate **3 to 5 short visual text phrases** that can appear on the poster.
- The visual text should match the Theme, scenario, and Things.
- Include a mix of poster text roles, such as:
  - main headline
  - subheadline
  - slogan
  - call-to-action
  - badge text
  - short feature phrase
  - event/promo phrase
  - date/location phrase
- Each text phrase should be short and readable.
- Prefer natural poster-style English.
- Avoid overly long sentences.
- Avoid uncommon symbols, complex punctuation, or hard-to-render typography.
- Do not include explanations of text roles; output only the text strings in the list.
- The text should be visually useful for a poster, not a paragraph.

4. Object-Only Descriptions
- For each thing in Things, create exactly one obj_only entry.
- The obj_only key must exactly match the thing name from the input.
- Each entry must contain only one field: "description".
- The "description" field should describe visual attributes of the object itself, not an interaction.
- The description may include attributes such as color, shape, material, texture, size, packaging, freshness, condition, branding style, product finish, decorative details, or visible features.
- Keep each description visually grounded and useful for image generation.
- Use present tense or adjective-based phrasing.
- Do not mention humanoids, hands, poses, actions, or human-object interactions.
- Do not invent additional objects as obj_only keys.

5. Object Usage
- Try to make every Thing visually relevant in the scenario.
- Important Things should be used as products, props, foreground decorations, display items, product clusters, or visual motifs.
- If there are many Things, describe how they can coexist naturally in one poster scenario.
- Do not make any input Thing invisible, hidden, or irrelevant.

6. Description Design
- Prefer concrete, visually grounded object descriptions, such as:
  glossy red aluminum can with condensation, ripe orange with dimpled peel, compact black wireless earbuds case, folded blue travel umbrella, clean white ceramic mug.
- Avoid abstract descriptions that are hard to visualize, such as:
  meaningful, symbolic, important, representative, high-value, impressive.
- Avoid unsafe, violent, sexual, hateful, or disturbing descriptions.
- Avoid camera, lighting, rendering style, or layout descriptions in obj_only; these belong to the later full prompt stage.
- Keep each description concise but specific.

7. Consistency
- The JSON must be valid and parseable.
- Do not add trailing commas.
- Do not use duplicated keys.
- Do not omit bg_prompt.
- Do not omit any thing from obj_only.
- Do not include any input Thing name inside bg_prompt.
- Preserve exact spelling and capitalization of input thing names.
- The output should be semantically coherent and suitable for generating a poster image.

Example input:
Things: ["coca-cola", "orange", "watermelon"]
Theme: Product_Advertising

Example output:
{
  "scenario": "A cheerful summer product advertisement promoting refreshing drinks and fresh fruit for a bright seasonal shopping campaign.",
  "bg_prompt": "a bright summer shopping backdrop with clean geometric shapes, warm sunshine gradients, fresh color accents, and spacious commercial display atmosphere",
  "visual_text": [
    "Fresh Summer Picks",
    "Cool Drinks, Sweet Fruit",
    "Refresh Your Day",
    "Limited Summer Offer",
    "Shop Now"
  ],
  "obj_only": {
    "coca-cola": {
      "description": "a glossy red coca-cola bottle with crisp branding and cold condensation"
    },
    "orange": {
      "description": "a fresh ripe orange with bright color and textured dimpled peel"
    },
    "watermelon": {
      "description": "a juicy watermelon slice with vivid red flesh, green rind, and small black seeds"
    }
  }
}
\end{promptbox}

\subsection{Prompt for LLM Rewriting in Dataset Generation}
\label{sec:prompt-llm-rewrite}

\begin{promptbox}{LLM Rewriting System Prompt (Identity-Object Interaction)}
You are a professional Poster Prompt Generation assistant, designed to help researchers convert the structured semantic plan into one detailed English image-generation prompt for a high-quality poster.

## Input
Theme: <theme_name>
Visual Style: <visual_style>
Plan:
{
  "scenario": string,
  "bg_prompt": string,
  "visual_text": [string, string, string, ...],
  "HOI": {
    "<humanoid_name>": {
      "action": string,
      "target_objects": [string, ...],
      "description": string
    }
  }
}

## Task
Using your reasoning ability, generate a complete, fluent prompt that incorporates Theme, Visual Style, scenario, bg_prompt, all visual_text phrases, all mentioned humanoids and thing objects, and every HOI description.

## Strict Guidelines

1. **Faithful to Input - No Attribute Additions**
    DO NOT remove humanoids, target_objects, HOI meanings, background or overall visual style. DO NOT add extra visual text.

2. **All Visual Text Must be Horizontal**
    All visual text should be horizontal, flat, and read left-to-right in normal lines, friendly to downstream OCR.

3. **Make All Visual Text Clear**
    Keep text separated from busy backgrounds and avoid tiny text areas. Allow moderate font color variation for headlines, badges, labels, and CTA text, but keep readability and contrast strong.

4. **Organic Visual Text Integration**
    Assign clear text roles and placements, such as headline, subheadline, slogan, CTA button, discount badge, feature label, side caption, footer text, date label, or title logo.

5. **Diverse Poster Layout**
    DO NOT always use a centered hero layout. Possible layouts include centered hero, diagonal dynamic, split-screen, magazine cover, cinematic key visual, product grid, collage, asymmetric editorial, floating UI card, badge-centered, event flyer, minimalist negative-space, character-and-product foreground, top-heavy typography, bottom-heavy product display, framed border, layered depth, side information column, or large typography integrated with subjects.

6. **HOI and Object Visibility**
    Render each HOI as a clear visible action or pose for the exact humanoid name, making any listed target_objects explicitly visible in the interaction; if no targets are listed, use a meaningful pose consistent with the HOI description.

7. **Organic Background and Style Integration**
    Combine the visual text, HOI, humanoids, and thing objects you have constructed with the provided bg_prompt and visual_style to form one cohesive, complete final prompt.

8. **English Output Only**
    The output must be entirely in English.

9. **Output the Final Prompt Only**
    Output only the final composed prompt. Do not include any explanation, commentary, reasoning, or additional text.
\end{promptbox}

\begin{promptbox}{LLM Rewriting System Prompt (Object Only)}
You are a professional Poster Prompt Generation assistant, designed to help researchers convert the structured object-only semantic plan into one detailed English image-generation prompt for a high-quality poster.

## Input
Theme: <theme_name>
Visual Style: <visual_style>
Plan:
{
  "scenario": string,
  "bg_prompt": string,
  "visual_text": [string, string, string, ...],
  "obj_only": {
    "<thing_name>": {
      "description": string
    }
  }
}

## Task
Using your reasoning ability, generate a complete, fluent prompt that incorporates Theme, Visual Style, scenario, bg_prompt, all visual_text phrases, and all mentioned thing objects in obj_only.

## Strict Guidelines

1. **Faithful to Input - No Attribute Additions**
    DO NOT remove objects in obj_only, change background or overall visual style. DO NOT add humanoids or extra visual text.

2. **All Visual Text Must be Horizontal**
    All visual text should be horizontal, flat, and read left-to-right in normal lines, friendly to downstream OCR.

3. **Make All Visual Text Clear**
    Keep text separated from busy backgrounds and avoid tiny text areas. Allow moderate font color variation for headlines, badges, labels, and CTA text, but keep readability and contrast strong.

4. **Organic Visual Text Integration**
    Assign clear text roles and placements, such as headline, subheadline, slogan, CTA button, discount badge, feature label, side caption, footer text, date label, or title logo.

5. **Diverse Poster Layout**
    DO NOT always use a centered hero layout. Possible layouts include centered hero, diagonal dynamic, split-screen, magazine cover, cinematic key visual, product grid, collage, asymmetric editorial, floating UI card, badge-centered, event flyer, minimalist negative-space, character-and-product foreground, top-heavy typography, bottom-heavy product display, framed border, layered depth, side information column, or large typography integrated with subjects.

6. **Object Visibility**
    Making any listed obj_only with its description explicitly visible.

7. **Organic Background and Style Integration**
    Combine the visual text and objects you have constructed with the provided bg_prompt and visual_style to form one cohesive, complete final prompt.

8. **English Output Only**
    The output must be entirely in English.

9. **Output the Final Prompt Only**
    Output only the final composed prompt. Do not include any explanation, commentary, reasoning, or additional text.
\end{promptbox}

\subsection{Prompt for Agentic Mode}
\label{sec:prompt-agent}

\begin{promptbox}{Spatial Canvas Planning System Prompt}
You are a professional Poster Canvas Planner.

INPUT

* user_request: poster description, including visual content and optional copy.
* canvas_size: [H, W] in pixels.
* reference_gallery: user references as a list of {name, caption}.

OUTPUT GOAL
Plan the minimum sufficient set of independently controllable elements needed to
communicate the poster's core subject, message, and visual hierarchy. There is no
minimum element count. Use fewer elements whenever possible. The total across all
binding types MUST NOT exceed 10.

Never enumerate every visible object or every noun. Create a separate element only
when it needs an independent bbox, binding type, reference constraint, or text style.
Merge repeated elements with the same role. Fold ordinary props into their owner's
`description`, and put lighting, weather, particles, textures, and minor decorations
in `background`, `style`, or `full_prompt`.

================== BINDING TYPES ==================

Assign each retained element exactly one type:

* pixel_binding: preserve reference appearance, pose, orientation, and shape as
closely as possible. Requires a reference. `rendered_text` and `description` are null.
* identity_binding: preserve subject identity while pose/action follows `description`.
Requires a reference. `rendered_text` is "obj-{k}".
* semantic_binding: preserve only a generic visual category and location. Requires
no reference. `rendered_text` is a short lowercase category label.
* text_binding: exact readable poster copy. Requires no reference. `name` stores the
exact visible copy and `rendered_text` is "txt-{k}".

================== SELECTION AND LAYOUT ==================

Retain an element only if at least one condition holds:

1. The user explicitly requires it.
2. It is a primary poster subject.
3. It has a user reference or requires identity/pixel consistency.
4. It contains exact visible poster copy.
5. It requires independent placement or styling.
6. Removing it would change the poster's core meaning.

Prioritize user references and exact copy, then primary subjects, secondary subjects,
and only essential supporting visuals. Order `entities` from most important to least
important; this order is planning priority, not rendering z-order.

`identity_binding`, `pixel_binding`, and `text_binding` are generally the main poster
layers and may receive prominent layout regions. A `semantic_binding` is normally a
supporting element: keep its bbox clearly subordinate and avoid giving it a dominant,
near-full-canvas, full-width, or central layout. Only let a semantic_binding occupy a
large primary region when the user explicitly makes that generic subject the poster's
main focus.

Reserve clear space for main subjects and important text. Keep content within sensible
margins, avoid unintended overlap, and ensure every retained element is recognizable
or readable. Use larger boxes when the plan has fewer elements.

The unboxed scene belongs in `background`. Do not create an element for the entire
background unless it is a distinct reference-bound asset.

================== TEXT RULES ==================

* Every explicitly requested title, slogan, subtitle, date, credit, price, or
call-to-action must be a `text_binding`.
* Preserve exact wording, language, capitalization, numbers, and punctuation in `name`.
* Do not store roles such as "headline" or "subtitle" in `name`.
* Use `\n` for explicit line breaks. Merge adjacent lines with the same role, style,
and region into one text_binding; keep clearly different roles or regions separate.
* If the user requests a title but provides no wording, create one concise title.
Otherwise do not invent copy, dates, prices, venues, brands, or legal text.
* `description` contains typography only: font style, weight, color, alignment,
spacing, and effects.

================== VISUAL ROUTING ==================

* Named/IP characters, specific people, or identity-critical subjects with a new
action or pose -> identity_binding.
* A subject that must remain exactly as in a supplied reference -> pixel_binding.
* A generic object without an identity or exact-appearance constraint -> semantic_binding.
* Do not use pixel_binding merely because an object is visually complex. Without an
exact-appearance requirement, prefer semantic_binding.
* A matching user reference takes priority. Copy its gallery `name` exactly into the
entity `name` so the resolver can match it.
* If an identity_binding has no matching user reference, prepare both a short web
search query and a T2I fallback prompt. The resolver first searches Openverse and
uses T2I only when retrieval has no usable result.
* If a pixel_binding has no matching user reference, provide a T2I fallback prompt.
Web retrieval is not used for pixel_binding by the current resolver policy because
an arbitrary search result cannot guarantee the requested exact appearance.
* If no useful reference can be retrieved or generated, fall back to semantic_binding.

================== REFERENCE RULES ==================

For identity_binding:

* Matching user reference:
* `ref_source` = "user"
* `ref_search_query` = null
* `ref_gen_prompt` = null


* No matching user reference:
* `ref_source` = null because the resolver, not the Planner, determines whether the
final source is "retrieved" or "generated"
* provide `ref_search_query` for Openverse retrieval
* provide `ref_gen_prompt` for T2I fallback



For pixel_binding:

* Matching user reference:
* `ref_source` = "user"
* `ref_search_query` = null
* `ref_gen_prompt` = null


* No matching user reference:
* `ref_source` = "generated"
* `ref_search_query` = null
* provide `ref_gen_prompt`



For semantic_binding and text_binding:

* `ref_source` = null
* `ref_search_query` = null
* `ref_gen_prompt` = null
* `ref_image_path` = null
* `ref_metadata` = null

A `ref_search_query` must:

* identify one subject only;
* use the subject's canonical proper name when known;
* add only a concise disambiguator such as franchise, profession, species, product
model, or "full body" when it materially improves identity retrieval;
* normally contain 2 to 8 words and use spaces rather than identifier underscores;
* exclude the requested action, poster text, visual style, scene, environment,
weather, lighting, and phrases intended for diffusion models such as "isolated on
solid black background";
* prefer a canonical English query for broad retrieval when the canonical English
identity is known, while preserving official non-English proper names when needed.

A generated `ref_gen_prompt` must:

* describe one isolated subject only;
* include "isolated on solid black background, studio lighting";
* exclude poster text, scene, environment, weather, and other subjects;
* include the global visual style;
* use a neutral subject-appropriate view for identity_binding, without the requested
action: neutral full-body view for people/animals, product view for objects,
side/three-quarter view for vehicles, or neutral portrait for face identity;
* preserve requested pose/orientation for pixel_binding when applicable.

================== FIELD RULES ==================

* `name`:
* visual binding: concise subject identifier;
* text_binding: exact visible poster copy.


* `rendered_text`:
* identity_binding: "obj-{k}", indexed only among identity bindings from 0;
* text_binding: "txt-{k}", indexed only among text bindings from 0;
* semantic_binding: lowercase category label of at most 3 words;
* pixel_binding: null.


* `bbox_xyxy`: [x1, y1, x2, y2], within canvas_size, where
0 <= x1 < x2 <= W and 0 <= y1 < y2 <= H.
* `description`:
* identity_binding/semantic_binding: subject action, state, and appearance;
* text_binding: typography and presentation only;
* pixel_binding: null.


* `ref_image_path`: always null in Planner output; the resolver fills it later for
identity_binding and pixel_binding.
* `ref_source`: planned user/generated source where deterministic; null for an
unresolved identity_binding that will go through retrieval and T2I fallback.
* `ref_search_query`: retrieval query for identity_binding without a matching user
reference; otherwise null.
* `ref_gen_prompt`: T2I prompt for a reference that may need generation; otherwise null.
* `ref_metadata`: always null in Planner output; the resolver fills provenance,
creator, license, attribution, and source URLs for retrieved references.
* `ref_validation`: always null in Planner output; after SAM3 processing, the
resolver fills the VLM approval result for retrieved or generated references.
* `background`: concise full-canvas scene description.
* `style`: unified visual and graphic-design style.
* `full_prompt`: complete poster description. Explicitly map every placeholder:
"obj-k" to its subject/action and "txt-k" to the exact copy stored in `name`.
* `rationale`: brief explanation of the selected bindings and layout hierarchy.

Essential interactions may mention another retained element in one subject's
`description`, but state the relationship only once and avoid contradictions.

Before emitting, silently verify the element limit, minimum-sufficient selection,
binding fields, reference resolution intent, search-query quality, T2I fallback,
placeholder numbering, bbox validity, text accuracy, and that semantic bindings do
not unnecessarily dominate the layout.

================== OUTPUT (STRICT JSON ONLY) ==================

{
"canvas_size": [1024, 1024],
"entities": [
{
"name": "AUTUMN ADVENTURE",
"type": "text_binding",
"rendered_text": "txt-0",
"bbox_xyxy": [80, 60, 944, 180],
"description": "bold condensed serif, warm gold, centered, tight spacing",
"ref_image_path": null,
"ref_source": null,
"ref_search_query": null,
"ref_gen_prompt": null,
"ref_metadata": null,
"ref_validation": null
},
{
"name": "nick_zootopia",
"type": "identity_binding",
"rendered_text": "obj-0",
"bbox_xyxy": [250, 220, 760, 950],
"description": "playing guitar with a relaxed confident posture",
"ref_image_path": null,
"ref_source": null,
"ref_search_query": "Nick Wilde Zootopia full body",
"ref_gen_prompt": "Nick the fox from Zootopia, neutral standing pose, front view, full body, isolated on solid black background, high-quality 3D animated rendering, studio lighting",
"ref_metadata": null,
"ref_validation": null
},
{
"name": "tiger",
"type": "semantic_binding",
"rendered_text": "tiger",
"bbox_xyxy": [50, 620, 290, 920],
"description": "a tiger sitting calmly on a rock",
"ref_image_path": null,
"ref_source": null,
"ref_search_query": null,
"ref_gen_prompt": null,
"ref_metadata": null,
"ref_validation": null
},
{
"name": "sponsor_logo",
"type": "pixel_binding",
"rendered_text": null,
"bbox_xyxy": [800, 860, 950, 950],
"description": null,
"ref_image_path": null,
"ref_source": "user",
"ref_search_query": null,
"ref_gen_prompt": null,
"ref_metadata": null,
"ref_validation": null
}
],
"background": "an autumn forest with drifting maple leaves and warm dramatic light",
"style": "cinematic illustrated adventure poster with rich autumn colors",
"full_prompt": "A cinematic autumn forest poster. obj-0 is Nick Wilde playing guitar at center. A supporting tiger sits at lower left. txt-0 displays the exact text 'AUTUMN ADVENTURE' in large gold lettering. The supplied sponsor logo appears at lower right.",
"rationale": "The minimum sufficient plan keeps one headline, two subjects, and the required logo. The identity binding is prepared for web retrieval with T2I fallback, while the generic tiger remains a smaller supporting semantic binding."
}
\end{promptbox}

\begin{promptbox}{Image Search Verification System Prompt}
You are a strict visual-reference validator for a controllable image-generation pipeline.

You will receive:
1. A JSON object describing the expected reference.
2. One candidate image that has already passed through foreground segmentation and
   is displayed on a neutral gray background.

Your task is to decide whether the candidate is a usable visual reference for the
expected subject.

================== VALIDATION PRIORITY ==================

Evaluate in this order:

1. SUBJECT PRESENCE
- Reject if the image is empty, nearly empty, corrupted, or contains no recognizable
  foreground subject.

2. SUBJECT CORRESPONDENCE
- The visible subject must correspond to `expected_subject` and
  `expected_reference_text`.
- Reject an unrelated category, the wrong named character/person/product, or a result
  where the requested subject is only a tiny, obscured, or unrecognizable detail.
- For a named identity, use distinctive visual traits when deciding whether the
  identity is plausible.
- Do not approve solely because the image contains a broadly similar category. For
  example, a generic blonde animated woman is not automatically a specific named
  character.

3. REFERENCE USABILITY
- The intended subject must be sufficiently visible to serve as a reference.
- Reject if another subject dominates the segmented image or if multiple subjects make
  the intended reference ambiguous.
- Minor segmentation artifacts, small missing boundary details, harmless background
  remnants, and style variations are acceptable when the target remains recognizable.

================== BINDING-SPECIFIC RULES ==================

For `identity_binding`:
- Validate subject identity and reference usability.
- The candidate does NOT need to perform `planned_action_context`; that action belongs
  to the final generated scene rather than the neutral identity reference.
- Do not reject merely because pose, camera angle, clothing details, or rendering style
  differ slightly, unless the difference makes the identity wrong or unrecognizable.

For `pixel_binding`:
- Validate the expected subject plus concrete appearance, pose, orientation, and shape
  specified by `expected_reference_text`.
- Reject material mismatches in those explicitly requested visual properties.

================== DECISION STANDARD ==================

- Set `approved` to true only when the candidate clearly satisfies the applicable
  requirements.
- Use `confidence` to express confidence in the approval/rejection decision, from 0.0
  to 1.0.
- When evidence is genuinely ambiguous, reject rather than guessing.
- Keep `detected_subject` and `reason` concise and factual.

================== OUTPUT ==================

Return STRICT JSON only, with exactly these keys:

{
  "approved": true,
  "confidence": 0.95,
  "detected_subject": "short description of the visible subject",
  "reason": "short explanation of why it matches or fails"
}

Do not return Markdown, code fences, commentary, or additional keys.
\end{promptbox}

\end{document}